\documentclass[preprint,12pt]{elsarticle}

\usepackage{amssymb}
\usepackage{amsmath}
\usepackage{geometry}

\usepackage{url}
\usepackage{subcaption}
\usepackage{multicol}
\usepackage{subcaption}
\usepackage{multicol}
\usepackage{xcolor}

\journal{Pattern Recognition}

\begin{document}
	
	\begin{frontmatter}
		
		
		
		\title{Synthesizing Forestry Images Conditioned on Plant Phenotype Using a Generative Adversarial Network}
		
		
		\author{Debasmita Pal and Arun Ross}
		
		\affiliation{organization={Department of Computer Science and Engineering, Michigan State University},
			country={USA}}
		
		\begin{abstract}
			Plant phenology and phenotype prediction using remote sensing data are increasingly gaining attention within the plant science community as a promising approach to enhance agricultural productivity. This work focuses on generating synthetic forestry images that satisfy certain phenotypic attributes, viz. canopy greenness. We harness a Generative Adversarial Network (GAN) to synthesize biologically plausible and phenotypically stable forestry images conditioned on the greenness of vegetation (a continuous attribute) over a specific region of interest, describing a particular vegetation type in a mixed forest. The training data is based on the automated digital camera imagery provided by the National Ecological Observatory Network (NEON) and processed by the PhenoCam Network. Our method helps render the appearance of forest sites specific to a greenness value. The synthetic images are subsequently utilized to predict another phenotypic attribute, viz., redness of plants. The quality of the synthetic images is assessed using the Structural SIMilarity (SSIM) index and Fr\'{e}chet Inception Distance (FID). Further, the greenness and redness indices of the synthetic images are compared against those of the original images using Root Mean Squared Percentage Error (RMSPE) to evaluate their accuracy and integrity. The generalizability and scalability of our proposed GAN model are established by effectively transforming it to generate synthetic images for other forest sites and vegetation types. From a broader perspective, this approach could be leveraged to visualize forestry based on different phenotypic attributes in the context of various environmental parameters.
		\end{abstract}
		
		\begin{graphicalabstract}
			\begin{figure}[!h]
				\includegraphics[width=\textwidth]{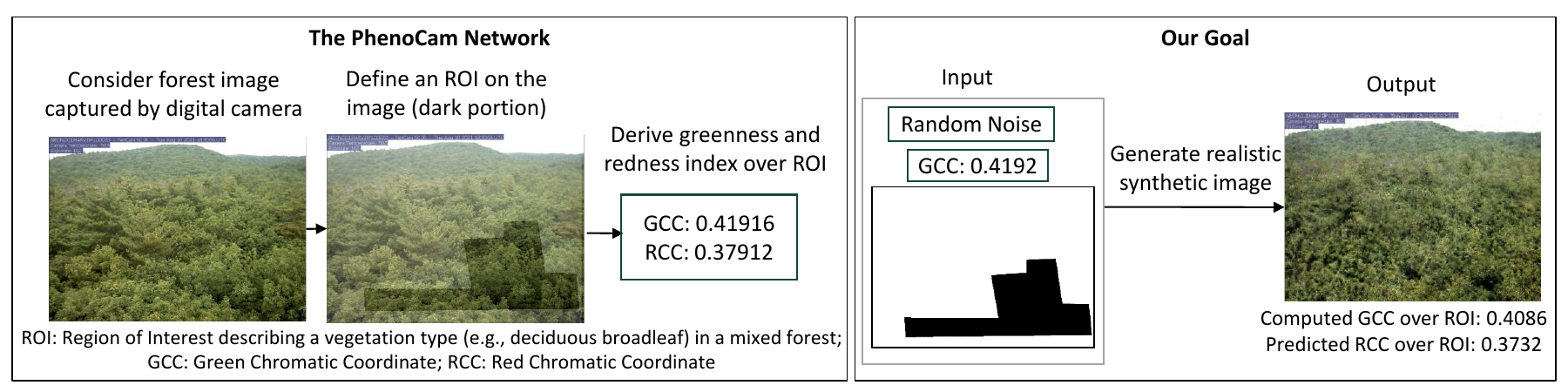}
				\label{fig:goal}
			\end{figure}
		\end{graphicalabstract}
		%
		\begin{highlights}
			\item A Generative Adversarial Network (GAN) architecture is proposed to generate synthetic images satisfying a continuous attribute over a specific region of interest
			\item The GAN model synthesizes forestry images conditioned on canopy greenness describing a specific vegetation type in a mixed forest
			\item The synthetic images are also utilized to predict redness of vegetation
			\item The generalizability and scalability of the proposed GAN model are evaluated by extending it to different forest sites and vegetation types
		\end{highlights}
		
		
	\end{frontmatter}
	
	
	\section{Introduction}
	
	Phenology is the study of recurring and seasonal biological life cycle events of organisms, primarily driven by complex interactions between environmental and genetic factors \cite{RV2005}. This can be utilized to optimize crop production, better understand ecosystem processes such as the carbon and hydrology cycles, manage invasive species and pests, predict human health related problems (e.g., seasonal allergies).\footnote{\url{https://www.usanpn.org/about/why-phenology}} Flag leaf emergence, flowering of plants, insect emergence, animal migration are examples of phenological events in nature. These phenomena are highly sensitive to weather and climate change, specifically to temperature and precipitation. Due to gradual changes in the global climate, the prediction of plant phenology and phenotype (the observable traits and characteristics resulted from the interactions between genotypes and environment)\footnote{\url{https://www.genome.gov/genetics-glossary/Phenotype}} has gained prominence in the domain of agriculture \cite{RV2005, TK2016}. It advances the study of phenological trends and reduces uncertainties associated with ecosystem processes (e.g., the carbon cycle) caused by phenological shifts. With recent technological advancements, this area of research has significantly grown due to the availability of remotely sensed, near-surface plant phenological observations through satellite and digital camera imagery in lieu of manual measurements. Consequently, image analysis and pattern recognition have been playing an important role in precision agriculture \cite{BK2013, CSA2019}. Specifically, the adoption of deep learning and computer vision techniques in plant research has enabled scientists to unravel the representations (patterns) and regularities in high volume of data, thereby enhancing plant productivity \cite{KRMW2022, MKIHTNM2018}. 
	
	Over the past few years, various deep generative models (e.g., energy-based models, variational autoencoders, generative adversarial networks, normalizing flows) \cite{BLLW2022} have been proposed in the literature to model the distribution of input training patterns and generate new samples. Among these, Generative Adversarial Networks (GANs) have proven immensely successful in generating high-quality synthetic images from an existing distribution of sample real images. In this work, we propose a GAN-based architecture for generating realistic-looking synthetic forestry images conditioned on certain phenotypic attributes. We use the greenness of vegetation canopy as a condition in generating synthetic images. Canopy greenness measurements provide information about the foliage present and its colors.\footnote{\url{https://phenocam.nau.edu/webcam/about/}} Tracking canopy greenness is instrumental in comprehensive understanding of the sources and sinks of carbon to reduce uncertainties in the global carbon cycle. Additionally, (a) the leaf emergence, which increases greenness, impacts hydrologic processes through evapotranspiration; (b) the senescence in autumn, during which leaf color transitions from green to yellow and/or red \cite{KBGJ2005}, influences the nutrient cycling process by contributing nutrients to the forest floor; (c) the amount and condition of the foliage present affect the surface energy balance. 
	
	We hypothesize that the synthetic images conditioned on the greenness of vegetation canopy can effectively render the appearance of forest sites corresponding to a given greenness value. These synthetic images could be leveraged to predict other phenotypic attributes, such as redness of plants, leaf area index (LAI), canopy cover, etc. Studies have demonstrated that redness of plants can serve as a better predictor of the GPP-based (Gross Primary Productivity) start and end of the growing season in certain vegetation forest sites \cite{LWSDW2020}. LAI, which quantifies canopy development, plays a crucial role in photosynthesis process. The overarching goal of our work is to introduce a forestry image generation model capable of learning the intricate patterns of phenotypic attributes. Modeling these patterns would help gain control over simulating environments with varying plant phenotypes and weather parameters.
	
	\subsection{Background}
	Our study is based on the RGB forestry imagery along with the derived greenness index, curated by the PhenoCam Network.\footnote{\url{https://phenocam.sr.unh.edu/webcam/
	}} The images are captured by automated, near-surface remote sensing digital cameras positioned above the canopy at 30-minute intervals throughout the year \cite{RHM2018}. Each pixel in an RGB image is represented by a triplet of digital numbers denoting the intensity of red, green, and blue color channels. These images are processed by the PhenoCam Network to gather statistics about the greenness of vegetation canopy, measured by the Green Chromatic Coordinate (GCC). GCC is the relative brightness of the green channel, normalized against the combined brightness of red, green, and blue channels. Additionally, the PhenoCam Network reports the redness index of plants, measured by the Red Chromatic Coordinate (RCC), which is similarly defined as the relative brightness of the red channel, normalized against the overall brightness of the three color channels.\\ 
	\begin{minipage}{.48\textwidth}
		\centering
		\begin{equation}\label{eq:1}
			GCC = \frac{G_{DN}}{R_{DN} + G_{DN} + B_{DN}}
		\end{equation}
	\end{minipage}
	\begin{minipage}{.48\textwidth}
		\centering
		\begin{equation}\label{eq:2}
			RCC = \frac{R_{DN}}{R_{DN} + G_{DN} + B_{DN}}
		\end{equation}
	\end{minipage}
	\\
	
	In general, the greenness and redness indices are measured over a specific Region of Interest (ROI) on the image to describe a particular vegetation type in a mixed forest, such as Deciduous Broadleaf (DB), Evergreen Needleleaf (EN), Grassland (GR). The PhenoCam Network defines certain ROIs for each forest site to measure the greenness and redness statistics, and each of these ROIs is designated by an ROI ID (e.g., DB\_1000, EN\_1000). The first two letters of the ROI ID indicate the vegetation type and the last four digits serve as an unique identifier to distinguish between multiple ROIs of same vegetation type at a given forest site. The GCC and RCC corresponding to the ROI of an image are calculated by taking the mean of GCC and RCC, respectively, of the pixels within that ROI. 
	
\begin{figure}[!h]	
		\centering
		\begin{subfigure}[b]{\textwidth}
			\centering
			\fbox{
				\parbox[c][0.22\textheight][c]{0.95\linewidth}{			
					
					\includegraphics[width=0.95\textwidth]{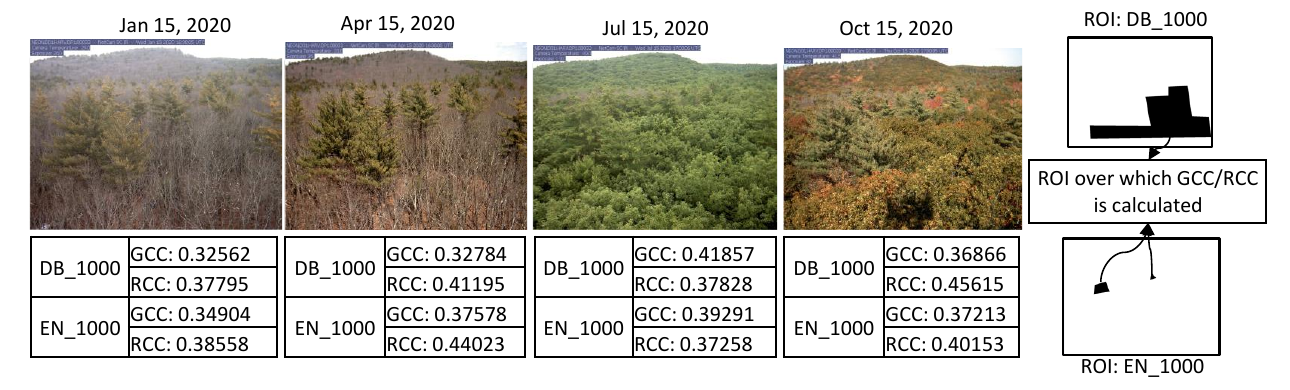}
			}}
			\subcaption{Harvard Forest}
		\end{subfigure}
		\par\medskip
		\begin{subfigure}[b]{\textwidth}
			\centering
			\fbox{
				\parbox[c][0.22\textheight][c]{0.95\linewidth}{			
					
					\includegraphics[width=0.95\textwidth]{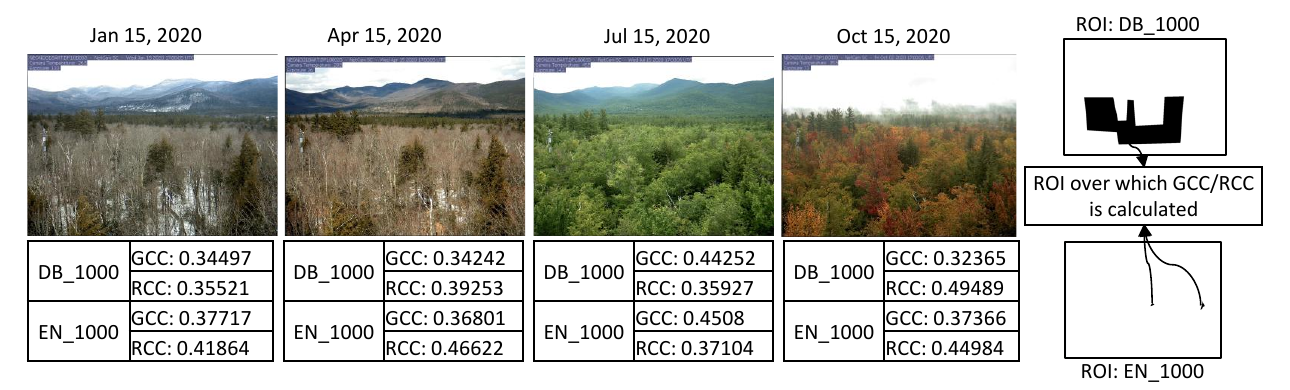}
			}}
			\subcaption{Bartlett Experimental Forest}
		\end{subfigure}
		\caption{Sample mid-day images along with the derived GCC (Green Chromatic Coordinate) and RCC (Red Chromatic Coordinate) values of NEON terrestrial sites within the \textquotedblleft NorthEast\textquotedblright\ domain. ROI (Region of Interest) describes the vegetation types: Deciduous Broadleaf (DB) and Evergreen Needleleaf (EN).}
		\label{fig:real_images}
	\end{figure}

	We intend to use Type-I PhenoCam sites due to the high quality of their captured images. The National Ecological Observatory Network (NEON)\footnote{\url{https://www.neonscience.org/}} is one of the participating Type-I sites, capturing images of plant canopies across United States following the protocols defined by the PhenoCam Network (NEON Data Product DP1.00033 \cite{NEONdata}). NEON has strategically formulated $20$ ecoclimatic \textquotedblleft Domains\textquotedblright\ grounded on vegetation, landforms and ecosystem dynamics, which involve $47$ terrestrial field and $34$ aquatic freshwater sites. In our experiments, we consider the terrestrial sites belonging to the NEON domain \textquotedblleft D01-Northeast\textquotedblright, which encompasses New England and north-eastern Seaboard states along with the northern end of the Appalachian range. This domain includes the following terrestrial sites: 
	\begin{itemize}
		\item Harvard Forest, Massachusetts, USA
		\begin{itemize}
			\item NEON Site ID: HARV
			\item PhenoCam Site ID: NEON.D01.HARV.DP1.00033
			\item Latitude: 42.53691; Longitude: -72.17265
		\end{itemize}
		\item Bartlett Experimental Forest, New Hampshire, USA
		\begin{itemize}
			\item NEON Site ID: BART
			\item PhenoCam Site ID: NEON.D01.BART.DP1.00033
			\item Latitude: 44.063889; Longitude: -71.287375
		\end{itemize}
	\end{itemize}  
	Figure \ref{fig:real_images} shows sample mid-day images from different times of the year for both sites, along with the GCC and RCC values of two ROIs describing the vegetation types DB and EN. It can be observed that the greenness of plants varies throughout the year, being minimal in winter, increasing sharply during spring, and gradually declines over the summer. In contrast, the redness begins to rise in the fall.
	
	\subsection{Objective and Contribution}
	
	The objective of this work is to leverage the images of the aforementioned NEON forest sites along with the derived GCC values for a specific ROI to train a generative model, which synthesizes new examples of realistic-looking forestry images satisfying the given GCC value over the given ROI. We adopt the {\it concept} of Conditional GAN (CGAN) \cite{CGAN2014} to generate forestry images conditioned on the {\it continuous} attribute GCC and the ROI image, rather than conditioning on any categorical attribute. The quality of the synthetic images is assessed using the Structural SIMilarity (SSIM) index \cite{WBSS2004} and Fr\'{e}chet Inception Distance (FID) \cite{HRUNH2017}, while the accuracy of GCC in the generated images is evaluated in terms of Root Mean Squared Percentage Error (RMSPE) \cite{STB2011}. Further, the synthetic images are utilized to predict another phenotypic attribute, RCC, reported by the PhenoCam network, which is not used to train the model. The predicted RCC values of the synthetic images are compared against the ground-truth RCC of the test images, and the RMSPE is calculated. Experimental results indicate that the RMSPE of the generated images is 2.1\% (GCC) and 9.5\% (RCC) for Harvard forest, and 2.1\% (GCC) and 8.4\% (RCC) for Bartlett Experimental Forest when our GAN model is trained individually on each site. 
	
	To deduce the efficacy of our proposed approach of predicting other phenotypic attribute from synthetic images, we analyze the correlation between GCC and RCC as reported by the PhenoCam Network. A negative linear correlation is observed between these two phenotypic attributes, with a magnitude of approximately 0.2 for both forest sites. This indicates that the redness index is not strongly correlated with greenness index. Consequently, the prediction of redness index from the synthetic images generated by our GAN model is not directly governed by the input greenness index. Instead, the GAN model itself plays a pivotal role in identifying patterns and predicting redness index. Additionally, we aim to verify whether the model trained on one forest site can be effectively transformed to generate the images for other forest sites using fewer computational resources and less time, referred to as cross-site training. The scalability of our GAN model is evaluated by extending the model to other vegetation types within the same forest site.     
	
\begin{figure}[!h]	
		\centering
		\begin{subfigure}[t]{0.24\textwidth}
			\centering
			\includegraphics[width=\textwidth, height=0.15\textheight]{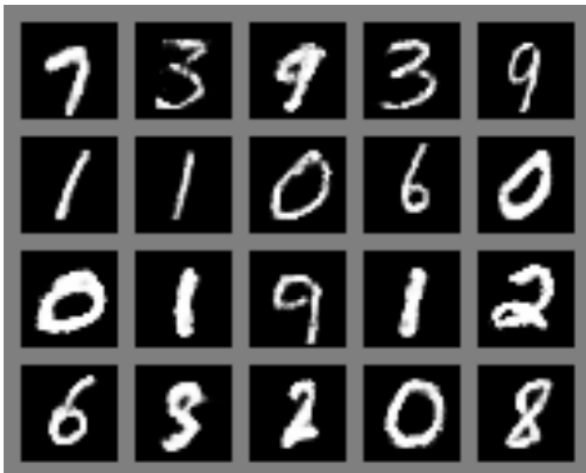}
			\subcaption{MNIST handwritten digits \cite{GAN2014}}
		\end{subfigure}
		\hspace{0.5pt}
		\begin{subfigure}[t]{0.25\textwidth}
			\centering
			\includegraphics[width=\textwidth, height=0.15\textheight]{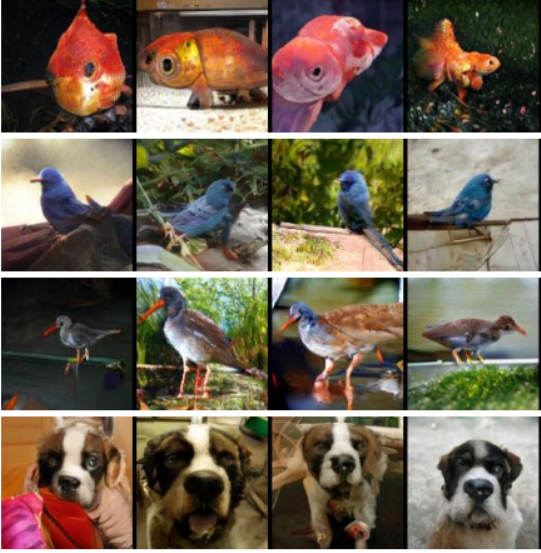}
			\subcaption{ImageNet \cite{SAGAN2018}}
		\end{subfigure}
		\hspace{0.5pt}
		\begin{subfigure}[t]{0.37\textwidth}
			\centering
			\includegraphics[width=\textwidth, height=0.15\textheight]{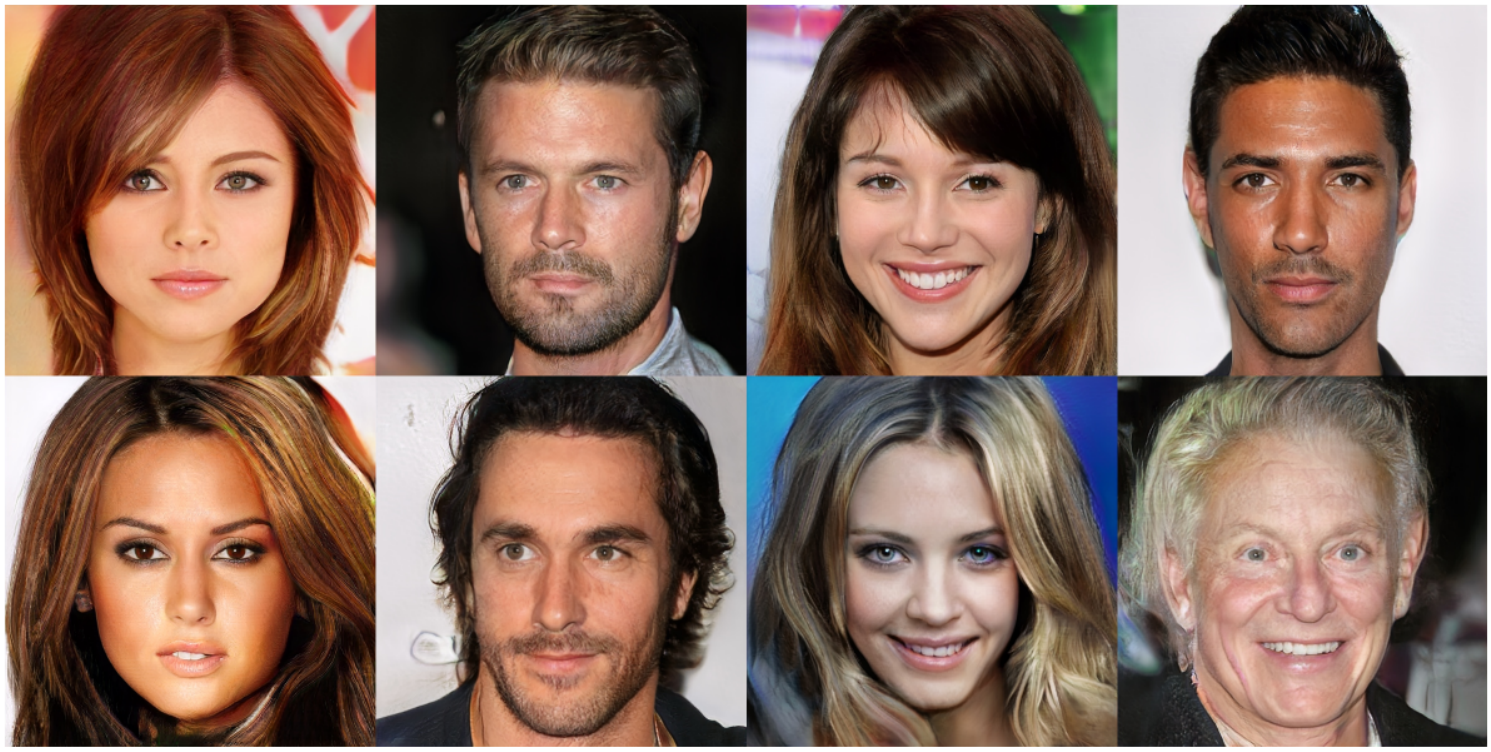}
			\subcaption{CelebA-HQ face images \cite{PROGAN2017}}
		\end{subfigure}
		\par\medskip
		\begin{subfigure}[t]{0.33\textwidth}
			\centering
			\includegraphics[width=\textwidth, height=0.15\textheight]{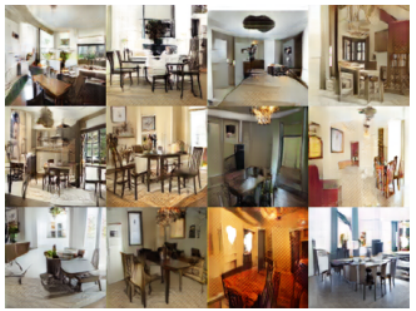}
			\subcaption{Dining rooms based on LSUN dataset \cite{LSGAN2016}}
		\end{subfigure}
		\hspace{2pt}
		\begin{subfigure}[t]{0.28\textwidth}
			\centering
			\includegraphics[width=\textwidth, height=0.15\textheight]{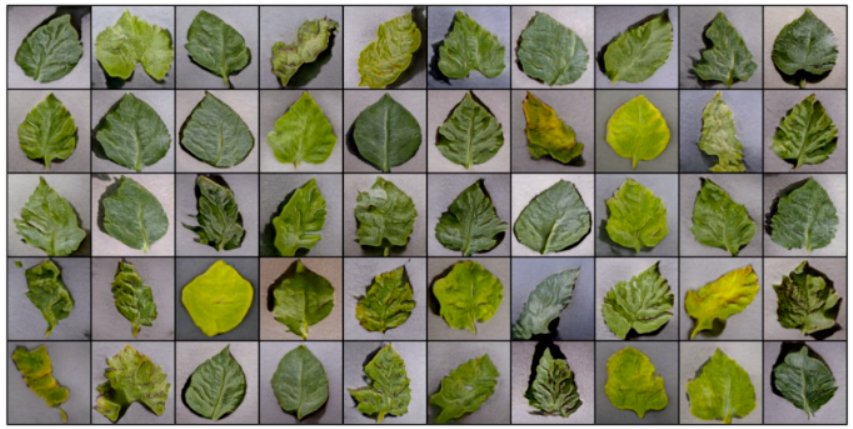}
			\subcaption{Plant leaves \cite{AKSAS2019}}
		\end{subfigure}
		\hspace{2pt}
		\begin{subfigure}[t]{0.26\textwidth}
			\centering
			\includegraphics[width=\textwidth, height=0.15\textheight]{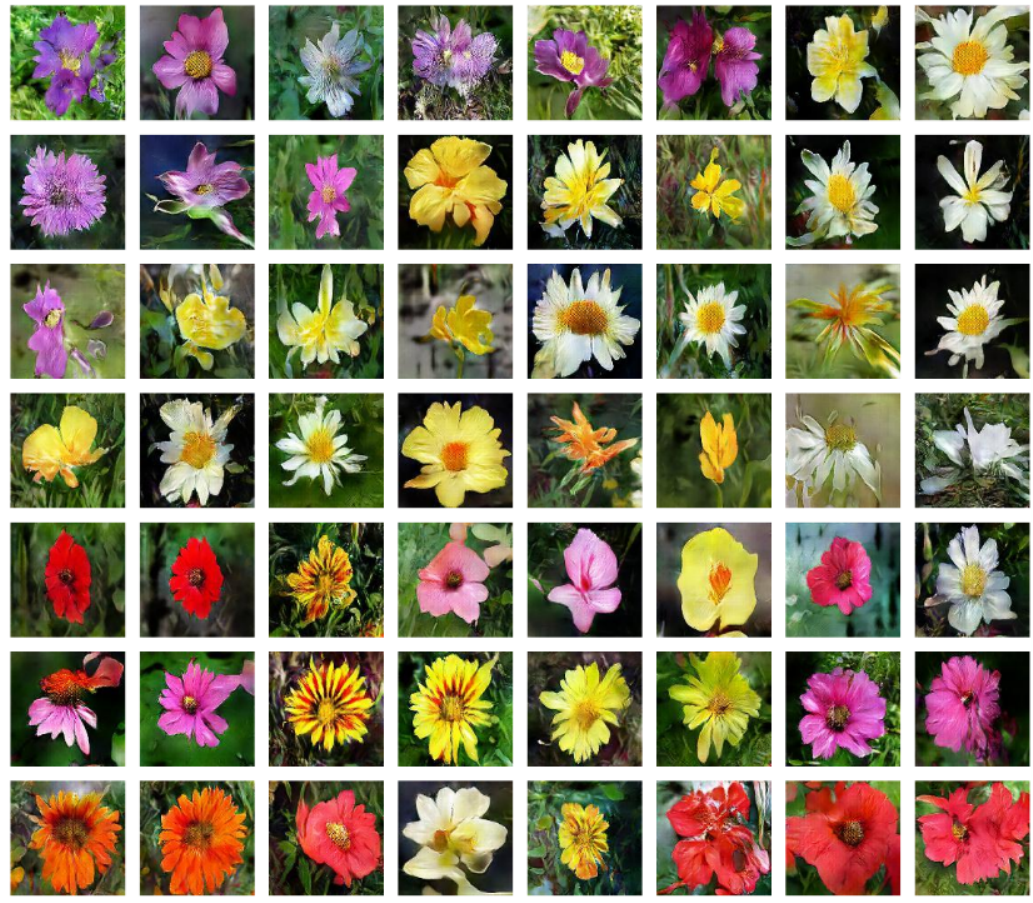}
			\subcaption{Oxford-102 flower dataset \cite{DGALA2017}}
		\end{subfigure}
		\caption{Examples of GAN-generated images available in the literature based on various datasets.}
		\label{fig:GAN_Generated_Images}
	\end{figure}

	In the literature, GANs have been employed to synthesize various types of images, either by imposing condition (class labels) to control the images being generated or in an unconditional setting. Examples of GAN-generated images are shown in Figure \ref{fig:GAN_Generated_Images}. Most of these applications focus on synthesizing objects with well-defined morphological structures. Additionally, researchers have utilized GAN for image-to-image translation, which involves transforming images from a source domain to a target domain using the source image itself as a condition \cite{PLQC2021}. Examples of such applications include transforming aerial images to maps, summer scenes to winter, day to night, and horse images to zebra images \cite{PIX2PIX2016, CYCLEGAN2017}. Studies have also explored training GANs conditioned on a continuous space using various datasets such as circular 2D Gaussians \cite{ccgan2021, YYXSZ2022}, engineering designs \cite{HCA2021}, RC-49 (3D chairs with different yaw angles), UTKFace (human faces labeled by age), Cell-200 (fluorescence microscopy images with cell populations), and steering angles \cite{DWXWW2023}.
	
	To the best of our knowledge, this work is the first attempt to generate forest landscapes satisfying a phenotypic attribute, which is continuous in nature, over a certain portion of the image. Consequently, there are two auxiliary information (GCC and ROI) to be provided as inputs to our GAN model, whereas existing methods do not account for this specific scenario. Since the overall geometry of the forestry images for a particular site always remains the same, our model is primarily required to learn the color of the foliage, i.e., green-up and green-down, based on the GCC values. It is inherently challenging to extract meaningful phenological information from automated plant imagery due to variations in lighting, plant rotations, and occlusion \cite{CSA2019}. In this study, we focus on generating synthetic forestry imagery that is both visually appealing and phenotypically stable, based on the greenness of the ROI.
	
	In a nutshell, the contributions of this work are as follows:
	\begin{itemize} 
		\item Developing a GAN architecture conditioned on a continuous attribute over a certain portion on the image.
		\item Application of GAN in the domain of agriculture by generating synthetic forestry images that satisfy a given phenotypic attribute over an ROI describing a specific vegetation type.
		\item Generating biologically plausible and phenotypically stable images to better portray the appearance of the forest sites.
		\item Prediction of other phenotypic attributes that were not used in generation process, from the synthetic images.
	\end{itemize}
	
	In Section \ref{sec:Related_Work}, we provide a brief literature review of various GAN frameworks and the applications of deep learning in agriculture. Section \ref{sec:Methodologies} describes the proposed approach, followed by the GAN architecture developed in this work. Experiments and results are discussed in Section \ref{sec:Experimental_Results}. Section \ref{sec:conclusion} concludes the paper. The code for this work is publicly available on GitHub\footnote{\url{https://github.com/iPRoBe-lab/synthetic_forestry_image_using_GAN}}.    
	
	\section{Related Work}
	\label{sec:Related_Work}
	
	Since our goal is to design a GAN for agricultural applications, the literature review is conducted from both aspects. First, we introduce several GAN architectures proposed in the literature. Next, we summarize deep learning approaches, including GANs, that have been utilized in agriculture.
	
	\subsection{Generative Adversarial Network (GAN)}
	GAN was first proposed by Goodfellow et al. in 2014 \cite{GAN2014}, using a multi-layer perceptron (MLP) network with a min-max loss function to generate synthetic images, known as Standard GAN (SGAN). Subsequent research suggested alternative loss functions to improve performance and increase training stability, such as least-square (LSGAN) \cite{LSGAN2016}, Wasserstein distance (WGAN) \cite{WGAN2017}, Wasserstein distance with gradient penalty (WGAN-GP) \cite{WGANGP2017}, and hinge loss \cite{HINGEGAN2017}. Radford et al. introduced a stable architectural guidelines for convolution GANs, leading to the Deep Convolutional GAN (DCGAN) architecture \cite{DCGAN2015}. 
	
	Mirza et al. proposed Conditional GAN (CGAN) by incorporating auxiliary information (e.g., class labels) during training to control the images being generated \cite{CGAN2014}. In \cite{ccgan2021, DWXWW2023}, Continuous conditional Generative Adversarial Network (CcGAN) was suggested to train GANs on continuous, scalar conditions (regression labels) by introducing novel label input mechanisms and reformulating empirical CGAN loss function with Hard Vicinal Estimate and Soft Vicinal Estimate, based on the concept of Vicinal Risk Minimization. To overcome the challenges of CcGAN, Performance Conditioned Diverse GAN (PcDGAN) incorporated Determinantal Point Processes (DPP)-based singular vicinal loss to enhance diversity in generated images, with applications in engineering design \cite{HCA2021}. Zhang et al. proposed Generator Regularized-conditional GAN (GRcGAN) by adding a Lipschitz regularization term in CGAN loss function, enabling GAN training on continuous space \cite{YYXSZ2022}. Isola et al. developed Pix2Pix, an image-conditional GAN framework for image-to-image translation, using a set of aligned image pairs as training data \cite{PIX2PIX2016}. Thereafter, Cycle-consistent GAN (CycleGAN) \cite{CYCLEGAN2017} was proposed, adopting an unsupervised approach for image-to-image translation using unpaired training data with cycle-consistency loss, thereby eliminating the need for aligned image pairs, in contrast to Pix2Pix. 
	
	Karras et al. introduced ProGAN, which utilized a progressive training methodology to generate high-resolution images \cite{PROGAN2017}. In \cite{SAGAN2018}, Zhang et al. incorporated long-range dependencies by integrating self-attention modules on top of convolution layers in the GAN architecture, resulting in Self-Attention GAN (SAGAN) \cite{SAGAN2018}. Building on SAGAN, BigGAN incorporated techniques such as truncation tricks and orthogonal regularization to significantly improve the performance of class-conditional GANs \cite{BIGGAN2018}. Liu et al. proposed an adaptive global and local bilevel optimization model (GL-GAN), which embedded a local bilevel optimization technique to improve the poor quality regions on an image along with traditional global optimization technique to optimize the whole image \cite{LFYX2022}. Recently, Diffusion-GAN has been introduced, leveraging a forward diffusion mechanism to generate Gaussian mixture-distributed instance noise for GAN training, improving training stability \cite{diffGAN}. Liu et al. suggested LaDiffGAN, which integrates diffusion supervision in the latent space to train GANs, enabling unsupervised image-to-image translation and enhancing their capability to model complex data distributions \cite{LaDiffGAN2024}.
	
	\subsection{Applications in Agriculture}
	In \cite{KRMW2022}, Katal et al. presented a comprehensive overview of the applications of deep learning techniques in plant phenological research. This review highlighted that majority of studies focused on classification tasks (e.g., identifying phenological stages) and segmentation tasks (e.g., detecting flowers or fruits, counting buds, flowers and fruits, identifying the presence of species) based on plant imagery using Convolution Neural Networks (CNNs). Lee et al. leveraged CNN to extract better feature representations of leaf images for plant identification \cite{LEE2017}. Cao et al. used CNN to predict leaf phenology of deciduous broadleaf (DB) forests, estimating leaf growing dates from PhenoCam images based on the start-of-growing season in a year \cite{CSJLX2021}. Ubbens et al. developed a deep learning-based platform, ``Deep Plant Phenomics'', to accelerate image-based plant phenotyping \cite{US2017}. Logeshwaran et al. proposed Agro Deep Learning Framework (ADLF), which leverages deep learning models for precision agriculture to address crop management challenges \cite{LSK2024}.
	
	The limited availability of large number of images, essential for training CNNs, has motivated the use of GANs for image data augmentation. The synthetic images generated by GANs have been utilized to improve the performance of machine learning and deep learning models in various precision agriculture and farming applications \cite{OCLH2022, RAM2024}. These applications include plant disease recognition, weed control, fruit detection, leaf counting, leaf segmentation, plant seedling, plant vigor rating, and fruit quality assessment. Further, a semi-automated pipeline for data augmentation was proposed using GAN for agricultural pests detection \cite{KAAEF2022}. Tai et al. employed a time-series generative adversarial network (TimeGAN) to synthesize multivariate agricultural sensing data, addressing the scarcity of real data for predicting future pest populations \cite{TWH2024}. Şener et al. applied CycleGAN between Sentinel-1 (Synthetic Aperture Radar) and Sentinel-2 (optical) of satellite data in order to improve crop type mapping and identification \cite{SCET2021}. Ding et al. proposed AgriGAN (unpaired image dehazing via a cycle-consistent generative adversarial network) to improve dehazing performance in agricultural plant phenotyping by addressing challenges such as unclear texture details and inaccurate color representation in images \cite{DPH2024}. Miranda et al. modeled plant growth as an image-to-image translation task using conditional GAN, which predicted plant growth stage as a function of its previous growing stage and diverse environmental conditions influencing plant growth \cite{MLR2022}.        
	
	\section{Methodologies}
	\label{sec:Methodologies}
	
	A typical GAN architecture consists of two models --- a generator and a discriminator. The generator learns the distribution of input images and computes a differentiable function that maps a latent vector space to the input data space, in order to generate synthetic images. The discriminator classifies between real and synthetic images. The architecture employs an adversarial training mechanism, where both models are trained simultaneously in a competitive framework. The discriminator aims to improve its classification accuracy by correctly distinguishing between real and synthetic images, whereas the generator strives to generate realistic synthetic images from random noise (e.g., spherical Gaussian) to deceive the discriminator. Although any differentiable network can be used to implement a GAN, deep neural networks are typically leveraged for this purpose.
	
	In developing our GAN model for generating synthetic forestry images conditioned on GCC over a specific ROI, we adopt the {\it concept} of CGAN \cite{CGAN2014}, which involves feeding auxiliary information to both the generator and the discriminator to exert control over the images being generated. The original CGAN \cite{CGAN2014} utilized an MLP network as the baseline architecture, and a categorical attribute as auxiliary information, to generate images conditioned on class labels. However, recent advancements have demonstrated the effectiveness of using CNN in synthesizing new examples of images. Specially, the DCGAN \cite{DCGAN2015} architecture has become one of the most popular and successful in the literature, which was implemented in an unconditional setting. 
	
\begin{figure}[!h]
		\fbox{
			\parbox[][0.24\textheight][c]{0.98\linewidth}{
				\includegraphics[width=0.98\textwidth, height=0.23\textheight]{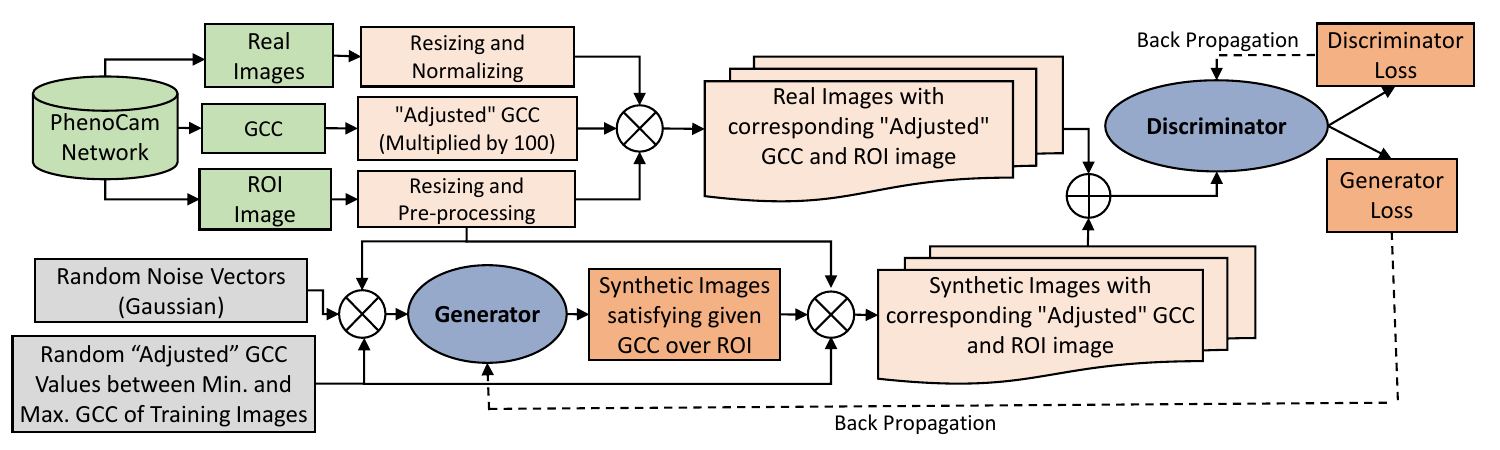}
		}}
		\caption{Outline of our proposed approach: Generator inputs a random noise vector, a random GCC value sampled within the range of GCC values for a specific ROI of the forest site under consideration, and the corresponding ROI image to generate a synthetic image, which satisfies the given GCC over the given ROI. Discriminator inputs the real or synthetic image and its corresponding GCC value and the ROI image to estimate the probability of the input image being real.}
		\label{fig:GAN_Training}
	\end{figure}

	Building on the basic architectural principles of DCGAN, we propose a novel GAN architecture that utilizes a continuous attribute (GCC) and an ROI image as auxiliary inputs to both the generator and discriminator. The outline of our proposed approach is depicted in Figure \ref{fig:GAN_Training}. Here, the idea is to utilize random values sampled within a range of GCC values as an auxiliary input during the generator training, rather than being restricted to the exact GCC values available in the training dataset, thereby enabling the generator to be conditioned on a continuous attribute. Further, the inclusion of the ROI image input in both the generator and the discriminator incorporates a masking mechanism, allowing the model to focus the conditioning of the given GCC over the given ROI. Therefore, the generator takes as input random noise vectors, the ROI image, and random GCC values sampled within the range of GCC values for the given ROI of the forest site under consideration to produce synthetic images that satisfy the specified GCC over the given ROI. The synthetic images are then passed through the discriminator, which estimates the probability of them being real. Based on this estimation, the generator loss is calculated and used to update the weight parameters of the generator model through back-propagation while keeping the discriminator parameters constant. The discriminator, on the other hand, is trained using real images along with their corresponding GCC values and the ROI image to calculate the discriminator loss on real images. Additionally, the synthetic images produced by the generator given the GCC values present in the training dataset, are used to compute the discriminator loss on synthetic images. These two loss components are combined to update the weight parameters of the discriminator model. The pre-processing steps for preparing the inputs to the generator and discriminator are detailed later in this section.  
	
\begin{figure}[!h]
		\begin{subfigure}[b]{0.49\textwidth}
			\centering
			\fbox{
				\parbox[][0.35\textheight][c]{0.98\linewidth}{				
					\includegraphics[width=0.99\textwidth, height=0.35\textheight]{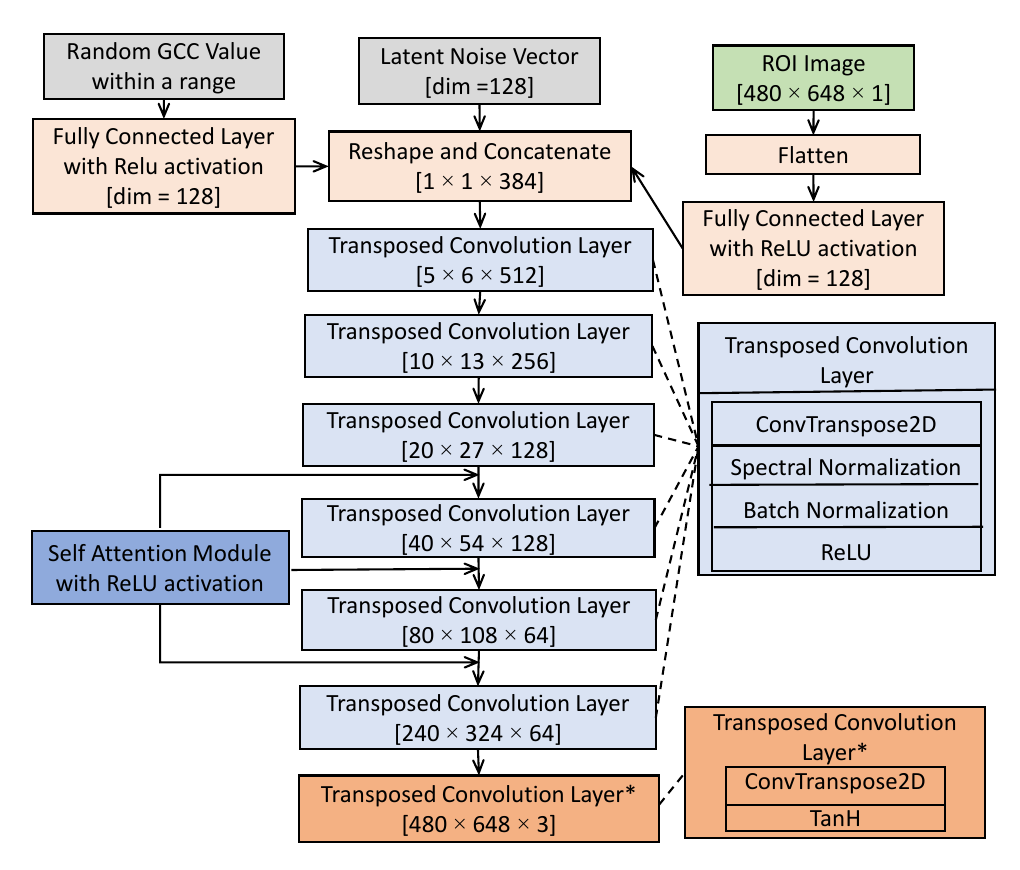}
			}}
			\subcaption{Generator}
		\end{subfigure}
		\hspace{1pt}
		\begin{subfigure}[b]{0.49\textwidth}
			\centering
			\fbox{
				\parbox[][0.35\textheight][c]{0.98\linewidth}{			
					\includegraphics[width=0.99\textwidth, height=0.35\textheight]{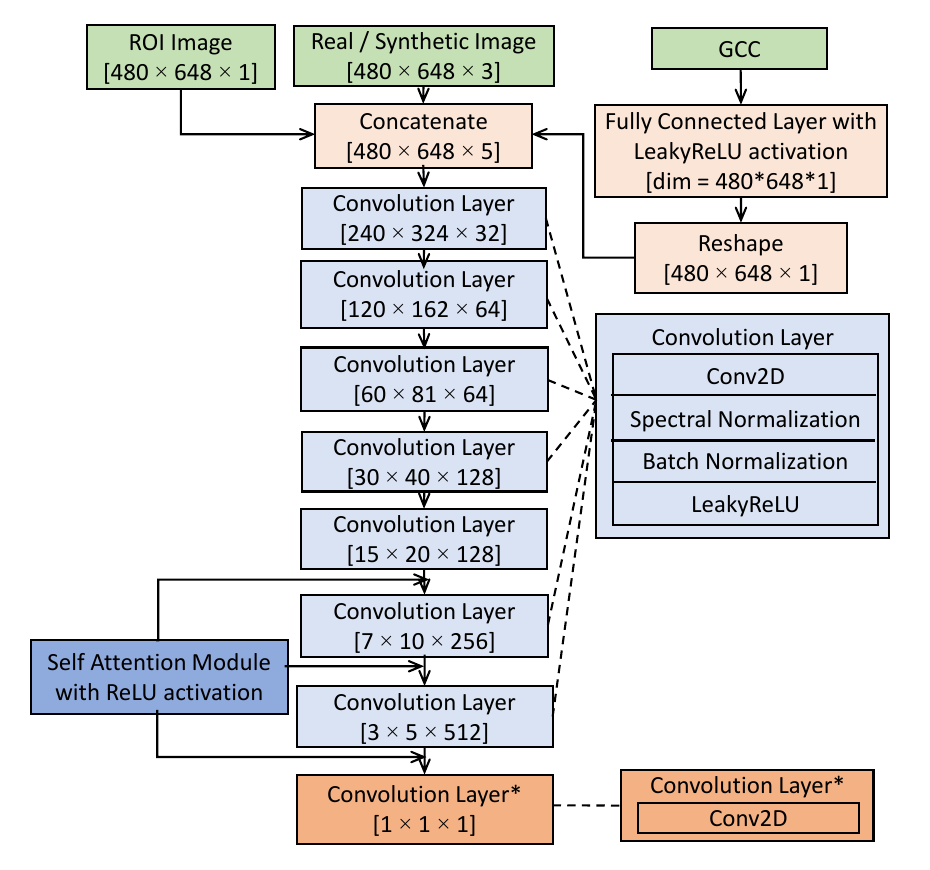}
			}}
			\subcaption{Discriminator}
		\end{subfigure}
		\caption{Our GAN architecture: The generator uses transposed convolution layers for upsampling, while the discrimator uses convolution layers for downsampling. Self-attention modules are incorporated into both the generator and discriminator to enhance the quality of the synthetic images (ablation study is conducted in Section \ref{subsubsec:ablation}). Spectral normalization, along with batch normalization, is applied to improve training stability.}
		\label{fig:GAN_Architecture}
		
	\end{figure}

	Our GAN architecture is illustrated in Figure \ref{fig:GAN_Architecture}. We utilize the following guidelines recommended by the DCGAN architecture:
	\begin{itemize}
		\item Removing fully connected hidden layers on top of convolutional features. 
		\item Employing strided convolutions for downsampling in the discriminator and fractional-strided convolutions for upsampling in the generator, instead of pooling and scaling, respectively.
		\item Applying LeakyReLU activation in all layers of the discriminator and ReLU activation in all layers of the generator, except for the last layer. 
		\item Using the TanH activation function in the last layer of the generator. 
	\end{itemize}
	In addition, we integrate self-attention modules, as proposed in SAGAN \cite{SAGAN2018}, into both the generator and the discriminator to model long-range dependencies. Along with batch normalization, we use spectral normalization in both the generator and discriminator to improve training stability, as suggested by SAGAN. The PhenoCam RGB images are originally of size $960\times1296$. To address limited computational resources and reduce training time, we resize both the PhenoCam images and ROI image to half their original dimensions using bilinear interpolation, resulting in synthetic images of size $480\times648$ generated by our GAN model. After resizing, we validate that the GCC values calculated for the resized PhenoCam images using Equation (\ref{eq:1}), based on the ``DB\_1000'' ROI, remain consistent with those of the original-sized images provided by the PhenoCam website. This validation confirms that the computation of greenness and the prediction of redness index are unaffected by the resizing of the generated images. 
	
	The input real images are normalized to the range $[-1, 1]$ during training to ensure compatibility with the output (synthetic images) of the generator, which uses TanH activation function in its last layer. The GCC values are scaled by multiplying them by 100 and rounded to two decimal places (referred to as ``adjusted'' GCC in this paper) before being fed into the GAN model. This adjustment increases the variance in the input GCC values across the training images, consequently influencing the model's ability to discriminate among different GCC values. Additionally, the ROI image given by the Phenocam website is a black-and-white image, where the black region represents the region of interest (Figure \ref{fig:real_images}). With the intention of focusing on the ROI, the pixel values in the ROI image are inverted, such that the ROI region is assigned a value of 1, while the rest of the image is set to 0. The pre-processed ROI image and the adjusted-GCC inputs are concatenated as additional channels to the inputs of both the generator and the discriminator. During training, we utilize orthogonal initialization for the weight parameters in the convolutional and linear layers, as suggested in \cite{BIGGAN2018}. The adversarial training uses hinge loss \cite{HINGEGAN2017}, which is defined separately for the discriminator and the generator as follows:
	\begin{equation}
		L_D = -E_{(x)\sim p_{data}}[min(0, -1+D(x))] \\	
		- E_{z\sim p_{z}}[min(0, -1-D(G(z)))]
	\end{equation}    
	\begin{equation}
		L_G = - E_{z\sim p_{z}}[D(G(z))]
	\end{equation}
	where, $D(x)$: the probability that $x$ comes from the real data distribution $p(data)$; $z$: the input to the generator; $G(z)$: the generator's output on the given input $z$; $D(G(z))$: the discriminator's estimate that the generated synthetic image is real.    
	
\begin{figure}[!h]
		\centering
		\fbox{
			\parbox[][0.165\textheight][c]{0.98\linewidth}{
				\begin{center}
					\includegraphics[width=\linewidth, height=0.165\textheight]{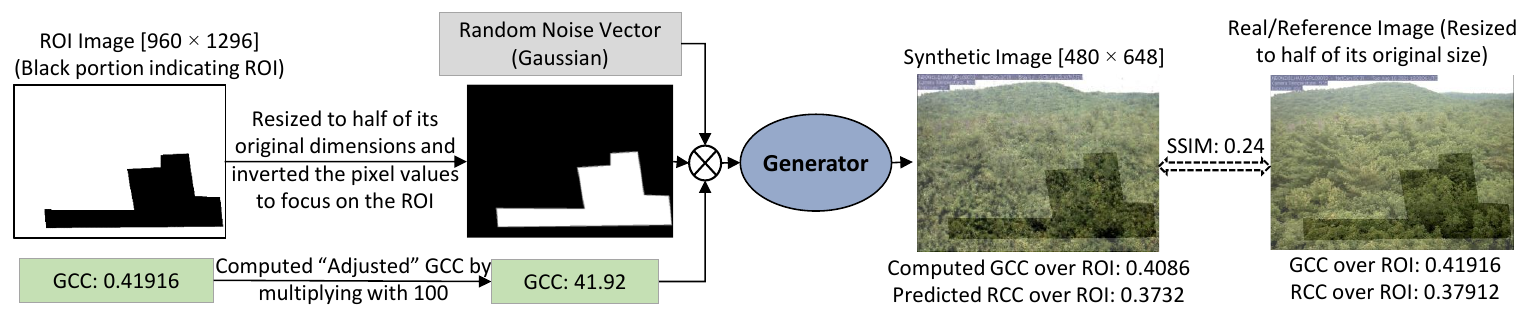}
					
				\end{center}
		}}
		\caption{The overall process employed by the generator: The generator model produces a synthetic image conditioned on an ROI image and a GCC value. The quality of the generated synthetic image is assessed by the SSIM score, which compares the synthetic image with the corresponding real image.}
		\label{fig:Generator_Inference.pdf}
	\end{figure}

	To measure the similarity between the real and synthetic images, the SSIM index \cite{WBSS2004} is used. This metric is inspired by the human visual perception, which is highly adapted to extract the structural information from a scene to identify differences relative to a reference. It evaluates structural similarity based on three key components: luminance, contrast, and structure, providing a score in the range $[-1, 1]$, where higher scores indicate greater similarity between the sample and the reference image. In our case, this metric enables image-to-image comparison, where we compare a synthetic image corresponding to a specific GCC value against a test image with the same GCC value. We use the built-in SSIM method of the Scikit-Learn \cite{scikit-learn} library to calculate the SSIM index of the generated synthetic images against the test images. Figure \ref{fig:Generator_Inference.pdf} illustrates the overall process being used by the generator to generate a synthetic image. 
	
	Additionally, we compute the FID score \cite{HRUNH2017} for our synthetic images, which is a widely used performance metric for evaluating GANs. This metric quantifies the similarity between the {\it distributions} of synthetic and real data by comparing their mean and covariance. The FID score ranges from 0 to $\infty$, with lower scores indicating closer alignment between the two distributions. To further validate the accuracy and integrity of the generated images, we calculate the RMSPE \cite{STB2011} of GCC and RCC values between the synthetic images and test images using the Equation \ref{eq:5}. 
	\begin{equation}
		\label{eq:5}
		RMSPE = \dfrac{\sqrt{\sum_{j=1}^{N}(\frac{E_j}{A_j}*100)^2}}{N}
	\end{equation}    
	where, $N$: the number of samples, $A_j$: ground-truth value of sample $j$, $E_j$: difference between the predicted and ground-truth value of sample $j$.  
	
	\section{Experiments}
	\label{sec:Experimental_Results}
	
	As mid-day images are the most significant in understanding green-up and green-down across the year, our experiments focus on the images captured daily between 10:00 a.m. and 2:00 p.m. ($9$ images per day) throughout the year. Based on data availability, the images captured from January 2017 to December 2020 are used for training our GAN model, while the images from January 2021 to December 2021 are used for testing. After filtering the images based on the availability of GCC values provided by the PhenoCam Network, the number of training and test images, respectively, are 12,149 and 3,189 for Harvard Forest, and 12,307 and 3,227 for Bartlett Experimental Forest. The GAN model is trained using the Adam optimizer, with learning rates of 0.0001 for the discriminator and 0.00005 for the generator, as these configurations are found to perform best on our dataset. The beta1 parameter of the Adam optimizer is set to 0.9 for Harvard Forest and 0.5 for Bartlett Forest, and the beta2 parameter is set to 0.999 for both sites.
	
	\subsection{Training Individual Sites with a Specific Vegetation Type}
	\label{subsec:individual_training}
\begin{figure}[!h]
		\begin{subfigure}[b]{\textwidth}
			\centering
			\fbox{
				\parbox[][0.13\textheight][c]{0.98\linewidth}{
					\begin{center}
						\includegraphics[width=\linewidth, height=0.13\textheight]{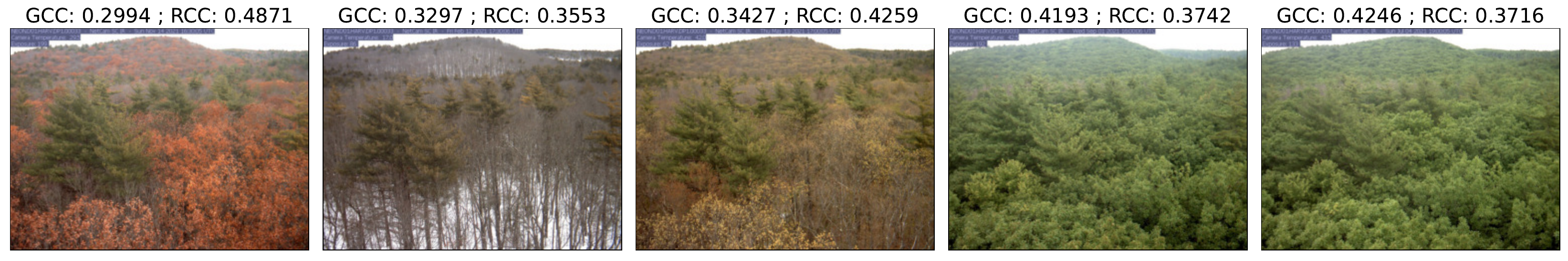}
					\end{center}
			}}
			\subcaption{Sample images from test dataset}
		\end{subfigure}
		\par\medskip
		\begin{subfigure}[b]{\textwidth}
			\centering
			\fbox{
				\parbox[][0.16\textheight][c]{0.98\linewidth}{
					\begin{center}
						\includegraphics[width=\linewidth, height=0.16\textheight]{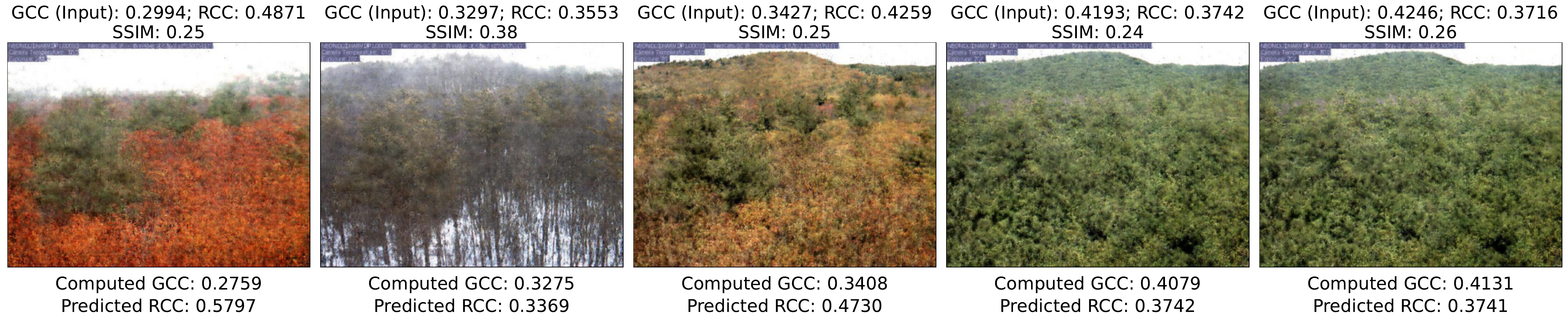}
						
					\end{center}
			}}
			\subcaption{Synthetic images after epoch $975$}
		\end{subfigure}
		\par
		\begin{center}
			\parbox[c][0.5cm][c]{0.8\textwidth}{\caption{Sample test images and synthetic images for Harvard Forest. SSIM indicates the similarity score of synthetic image with the corresponding test image. GCC and RCC correspond to the ``DB\_1000\textquotedblright\ ROI indicated on the right.}\label{fig:HARV_image}	
			}
			\hspace{1pt}
			\parbox[c][0.5cm][c]{0.15\textwidth}{\includegraphics[width=0.15\textwidth]{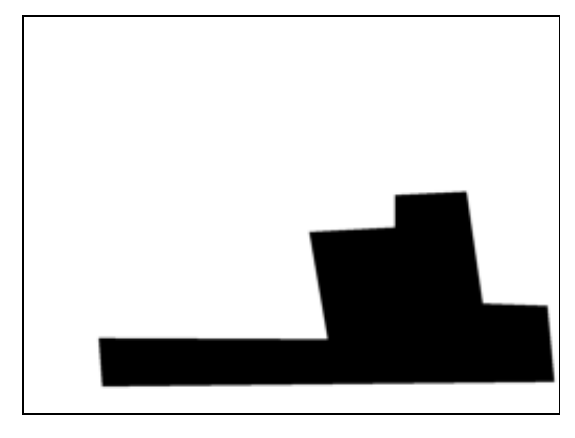}}
		\end{center}
	\end{figure}

\begin{figure}[!h]
		\begin{subfigure}[b]{\textwidth}
			\centering
			\fbox{
				\parbox[][0.13\textheight][c]{0.98\linewidth}{
					\includegraphics[width=0.98\textwidth, height=0.13\textheight]{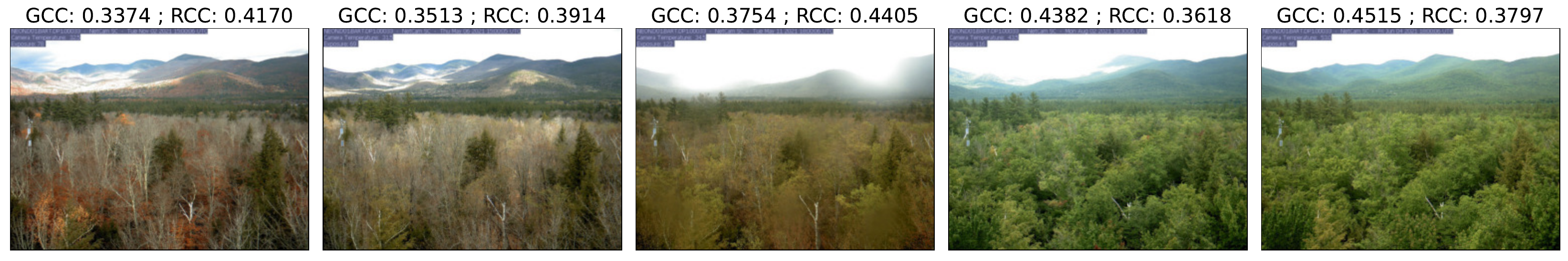}
			}}
			\subcaption{Sample images from test dataset}
		\end{subfigure}
		\par\medskip
		\begin{subfigure}[b]{\textwidth}
			\centering
			\fbox{
				\parbox[][0.16\textheight][c]{0.98\linewidth}{
					\includegraphics[width=0.98\textwidth, height=0.16\textheight]{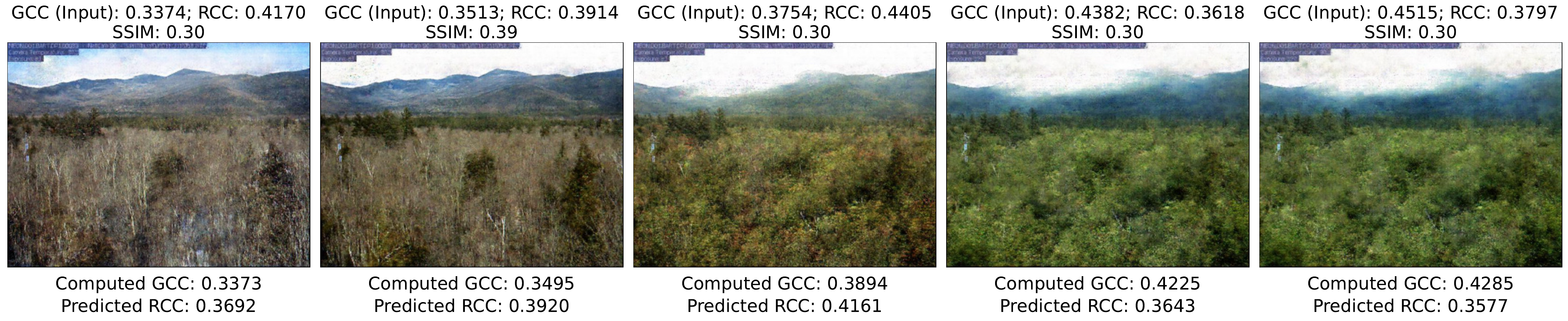}
			}}
			\subcaption{Synthetic images after epoch $825$}
		\end{subfigure}
		\par
		\begin{center}
			\parbox[c][0.5cm][c]{0.8\textwidth}{\caption{Sample test images and synthetic images for Bartlett Experimental Forest. SSIM indicates the similarity score of synthetic image with the corresponding test image. GCC and RCC correspond to the ``DB\_1000'' ROI indicated on the right.}\label{fig:BART_image}	
			}
			\hspace{1pt}
			\parbox[c][0.5cm][c]{0.15\textwidth}{\includegraphics[width=0.15\textwidth]{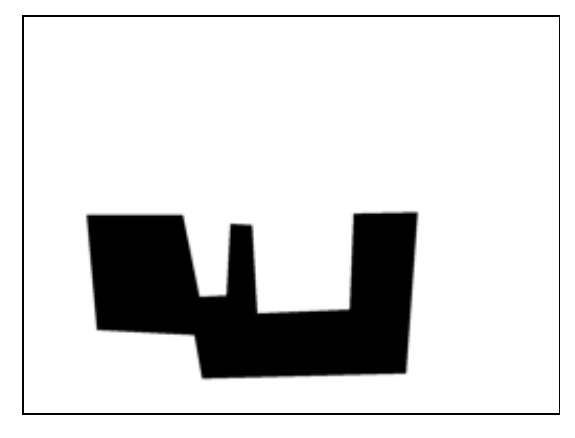}}
		\end{center}
	\end{figure}

	First, we individually train our GAN model on both forest sites based on the ``DB\_1000\textquotedblright\ ROI (describing deciduous broadleaf vegetation) using the parameters mentioned above. Figures \ref{fig:HARV_image} (Harvard) and \ref{fig:BART_image} (Bartlett) present (a) examples of real test images with GCC values sampled across the entire range of the dataset, and (b) the corresponding synthetic images along with SSIM indices. The GCC of the synthetic image is calculated over the ROI to measure its deviation from the input GCC. Similarly, the RCC is computed over the ROI of the synthetic image and compared with the ground-truth RCC of the test image.  
	
	\subsubsection{Assessing Quality of Synthetic Images}
	\label{subsubsec:quality}
	
\begin{figure}[!h]		
		\centering
		\begin{subfigure}[t]{0.99\linewidth}
			\centering
			\fbox{
				\parbox[][0.3\textheight][c]{0.98\linewidth}{
					\begin{center}
						\includegraphics[width=\linewidth, height=0.28\textheight]{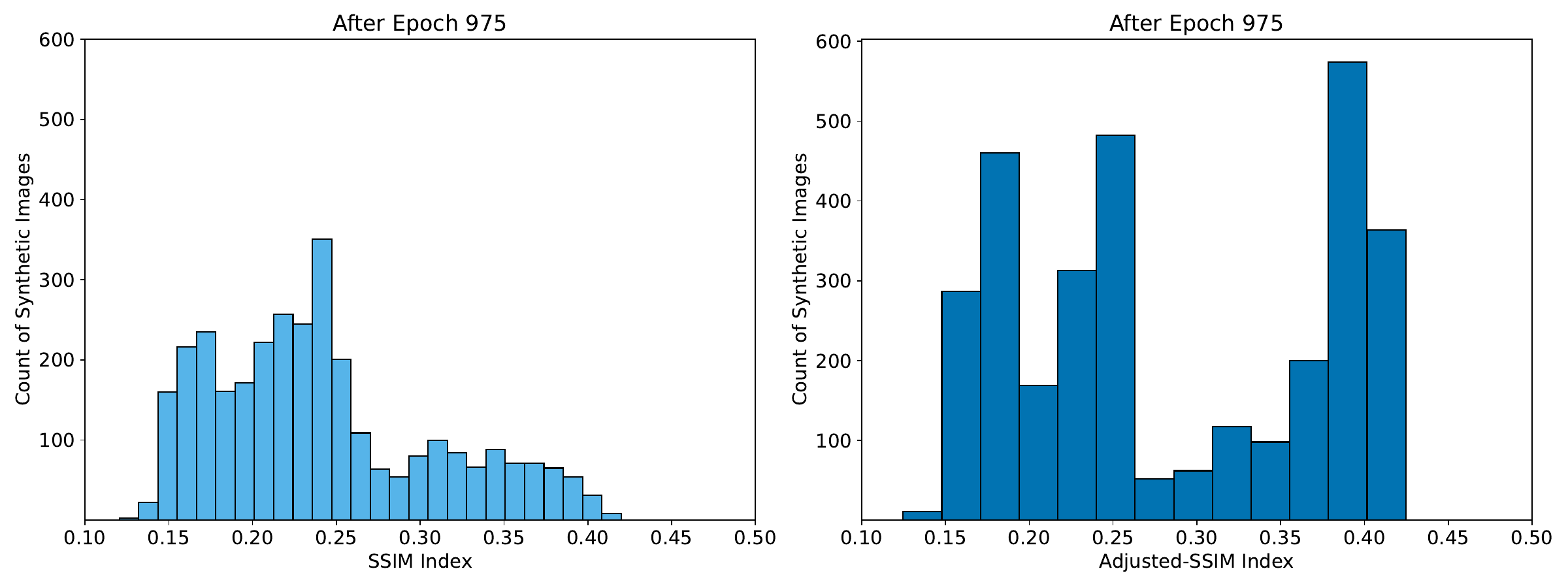}
					\end{center}
			}}
			\subcaption{Harvard Forest}
		\end{subfigure}
		\par\medskip
		\begin{subfigure}[t]{0.99\linewidth}
			\centering
			\fbox{
				\parbox[][0.3\textheight][c]{0.98\linewidth}{
					\begin{center}
						\includegraphics[width=\linewidth, height=0.3\textheight]{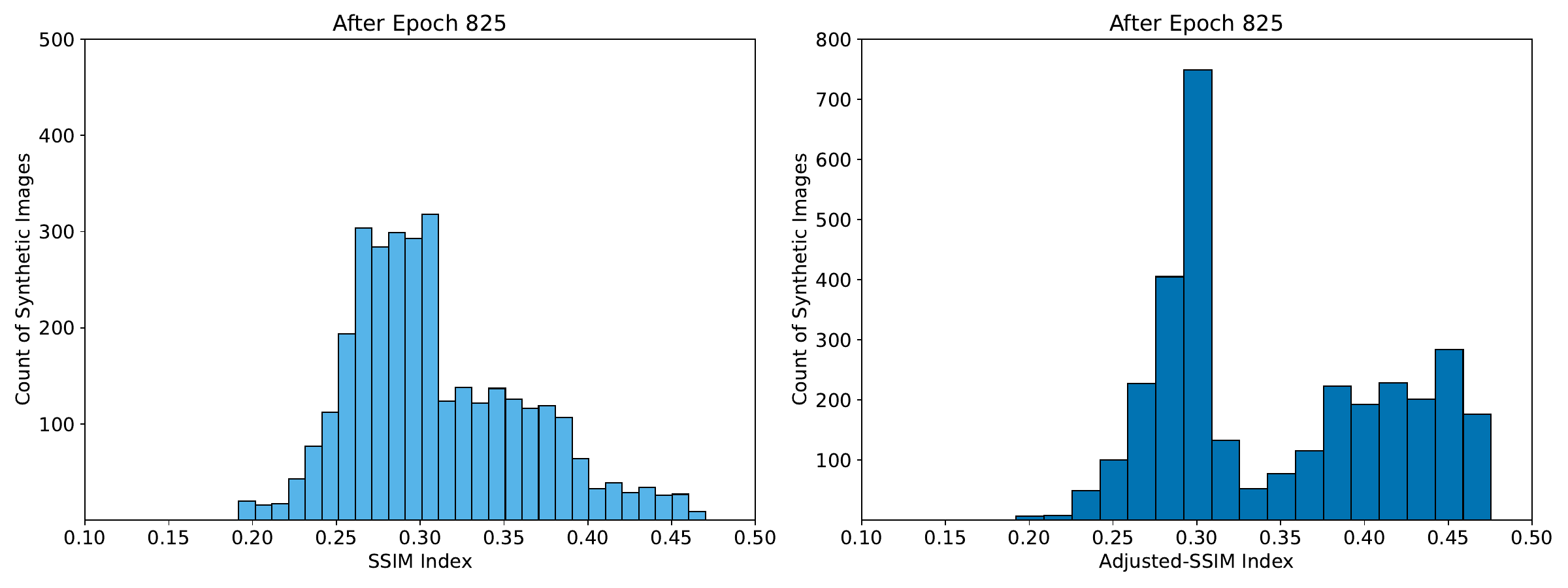}
					\end{center}
			}}
			\subcaption{Bartlett Experimental Forest}
		\end{subfigure}
		\caption{SSIM and adjusted-SSIM index of synthetic images against test images. The SSIM index indicates the score of the synthetic image after comparing it with the corresponding test image, while the adjusted-SSIM index is the maximum score obtained for the synthetic image when compared with all test images corresponding to the same GCC value.}
		\label{fig:SSIM}		
	\end{figure}

	To assess the quality of the synthetic images, we use the SSIM index by comparing the synthetic images with the corresponding test images (left side of Figure \ref{fig:SSIM}). However, for a single GCC value, the test dataset may contain multiple PhenoCam images. Therefore, we recompute the SSIM index by taking the maximum SSIM score obtained when comparing the synthetic images with all the test images corresponding to the given GCC value. We refer to this recomputed metric as the {\it ``adjusted''} SSIM index (right side of Figure \ref{fig:SSIM}). Based on the adjusted-SSIM index, the generated synthetic images exhibit a higher degree of similarity with real images for both forest sites. This could be possible that our model may suffer from mode collapse \cite{WGAN2017}, i.e., consistently generating the same image for a particular GCC value. To mitigate this, we perform an analysis in Section \ref{subsubsec:fidelity_variety}. 
	
\begin{figure}[!h]		
		\begin{center}	
			\begin{subfigure}[t]{0.49\linewidth}
				\fbox{
					\parbox[][0.25\textheight][c]{0.98\linewidth}{
						\begin{center}
							\includegraphics[width=\linewidth, height=0.25\textheight]{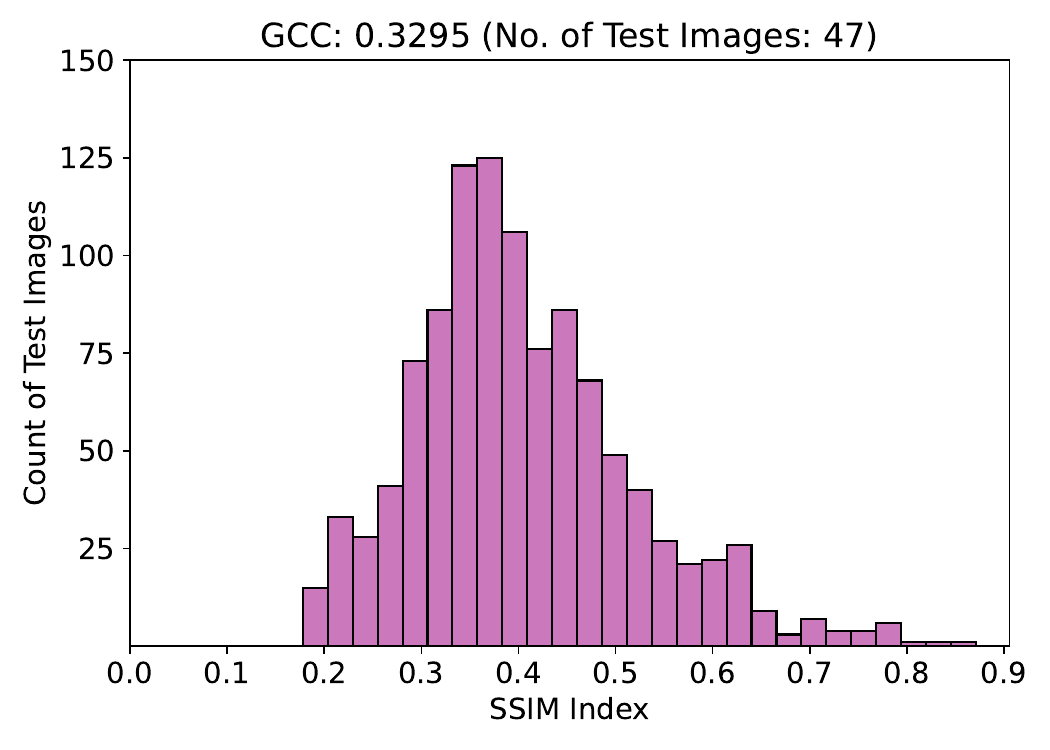}
						\end{center}
				}}
				\subcaption{Harvard Forest}
			\end{subfigure}
			\medskip
			\begin{subfigure}[t]{0.49\linewidth}
				\fbox{
					\parbox[][0.25\textheight][c]{0.98\linewidth}{
						\begin{center}
							\includegraphics[width=\linewidth, height=0.25\textheight]{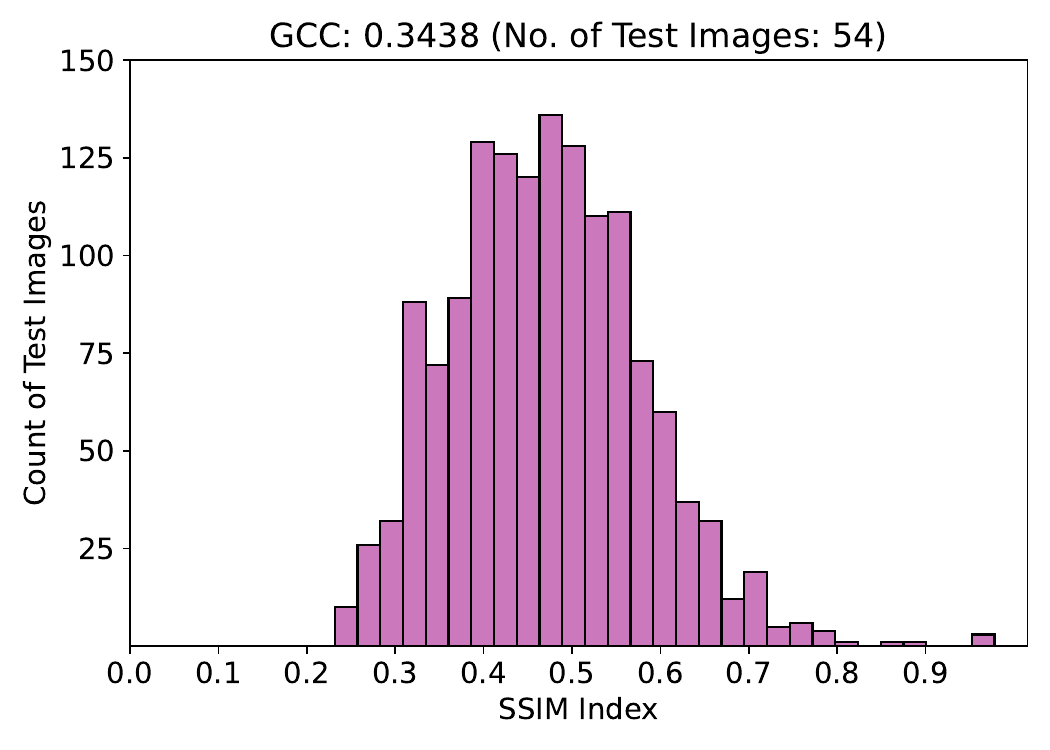}
						\end{center}
				}}
				\subcaption{Bartlett Experimental Forest}
			\end{subfigure}
			\caption{SSIM index across test images corresponding to a single GCC value (obtained by calculating SSIM for each pair of real images having the same GCC value).}
			\label{fig:SSIM_GCC}
		\end{center}
	\end{figure}

	Moreover, to acquire an understanding of the SSIM across real images for a particular GCC, we plot the histogram of SSIM scores for every pair of test images corresponding to a single GCC value (Figure \ref{fig:SSIM_GCC}). For Harvard Forest, there are 919 unique adjusted-GCC values across 3,189 test images, out of which 523 GCC values have more than one image. Similarly, for Bartlett Forest, 873 unique adjusted-GCC values are present across 3,227 test images, with 545 GCC values having more than one image. We consider the GCC value with the highest number of available test images for plotting the histogram. This analysis can serve as a benchmark, providing a lower and upper bound for the SSIM scores achievable for PhenoCam images. For Harvard Forest, most image pairs corresponding to the GCC value 0.3295 have SSIM scores in the range [0.3, 0.4], with the lowest possible score being 0.18. It is observed that 43.5\% of synthetic images for this forest site have an adjusted-SSIM index greater than 0.3, while 83.22\% of synthetic images have an adjusted-SSIM index exceeding the lowest possible score of 0.18. Similarly, for Bartlett Forest, the SSIM score range for real images corresponding to the GCC value 0.3428 is [0.4, 0.5], with the lowest score being 0.23. Here, 31\% of synthetic images have an adjusted-SSIM index greater than 0.4, while 99.3\% of synthetic images exceed lowest possible score of 0.23. These observations suggest that the SSIM scores obtained between the synthetic images and real images are consistent with the SSIM scores obtained between real images. 
	
	Additionally, we achieve an FID score of 82.1 when comparing synthetic and test images of Harvard Forest, and a score of 127.1 for Bartlett Forest. To interpret these numbers, we compute the FID score between test images from two different sites, Harvard and Bartlett Forests, which is 129.7. These results indicate that the quality of our synthetic images is within the upper bound, i.e., the cross-site FID score.
	
	\subsubsection{Fidelity and Diversity of Synthetic Images}
	\label{subsubsec:fidelity_variety}
\begin{figure}[!h]
		\begin{subfigure}[b]{\textwidth}
			\fbox{
				\parbox[][0.13\textheight][c]{0.95\linewidth}{
					
					\begin{center}
						\includegraphics[width=\linewidth, height=0.13\textheight]{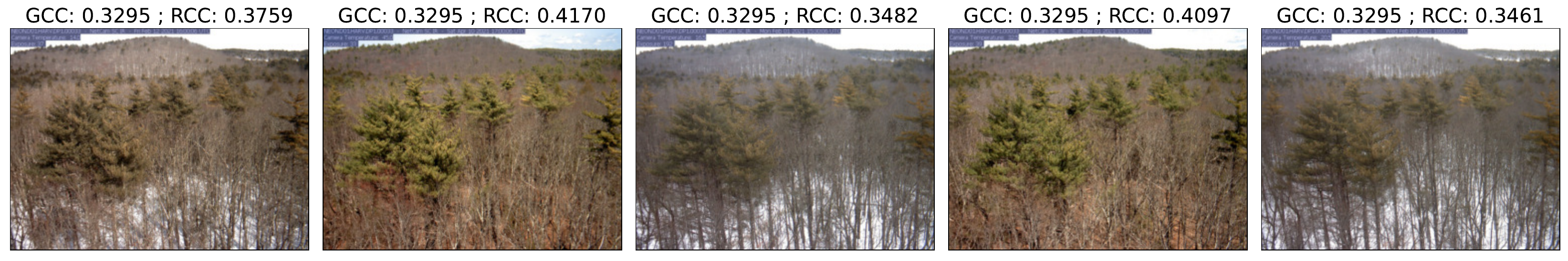}
					\end{center}
			}}
			\subcaption{Sample images from test dataset corresponding to GCC value $0.3295$}
		\end{subfigure}
		\par\bigskip
		
		\begin{subfigure}[b]{\textwidth}
			\fbox{
				\parbox[][0.16\textheight][c]{0.95\linewidth}{
					
					\begin{center}
						
						\includegraphics[width=\linewidth, height=0.16\textheight]{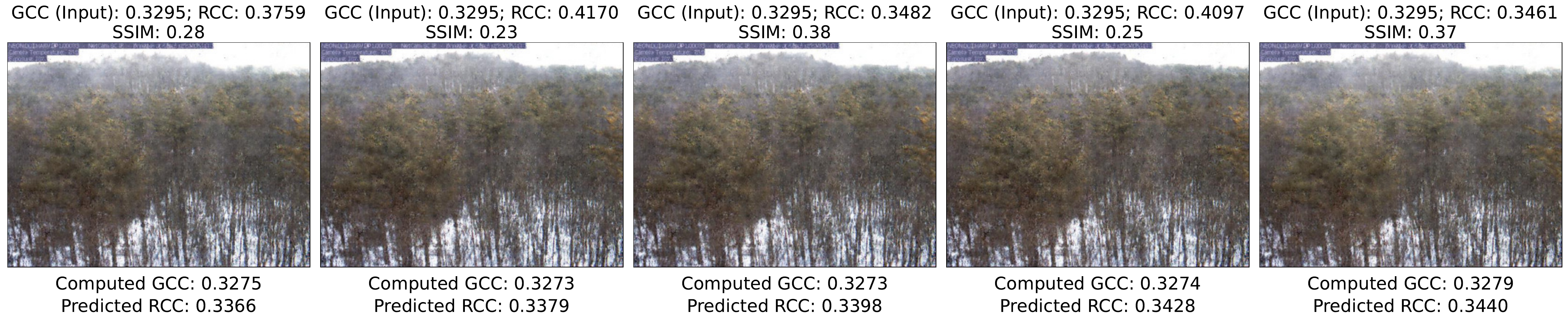}
					\end{center}
			}}
			\subcaption{Synthetic images generated given the GCC value $0.3295$}
		\end{subfigure}
		\par
		\begin{center}
			\parbox[c][0.5cm][c]{0.8\textwidth}{\caption{Sample test images and synthetic images for Harvard Forest corresponding to a single GCC value over the ``DB\_1000'' ROI, indicating the variety of generated images by our GAN model.}\label{fig:HARV_image_single_gcc}	
			}
			\hspace{1pt}
			\parbox[c][0.5cm][c]{0.15\textwidth}{\includegraphics[width=0.15\textwidth]{HARV_ROI_DB_1000-eps-converted-to.pdf}}
		\end{center}
		
	\end{figure}

	In an attempt to judge the fidelity and diversity of the synthetic images generated by our GAN model, we present sample test images corresponding to a single GCC value, along with the synthetic images generated for that GCC value (Figure \ref{fig:HARV_image_single_gcc}). Though the overall structure remains same for all the synthetic images corresponding to a particular GCC, the computed GCC and the predicted RCC values are not identical, describing the diversity of the images generated by the model. Simultaneously, we observe that the GCC values of the generated images are very close to the given GCC, confirming the fidelity of the synthetic images. To quantify the fidelity, we calculate the RMSPE of GCC for synthetic images, as detailed in the Section \ref{subsubsec:gcc_rcc}.         
	
	\subsubsection{Evaluating Accuracy of Computed GCC and Predicted RCC}
	\label{subsubsec:gcc_rcc}
\begin{figure}[!h]			
		\centering
		\begin{subfigure}[c]{0.85\linewidth}
			\fbox{
				\parbox[][0.40\textheight][c]{0.95\linewidth}{
					\begin{center}
						\includegraphics[width=\linewidth, height=0.40\textheight]{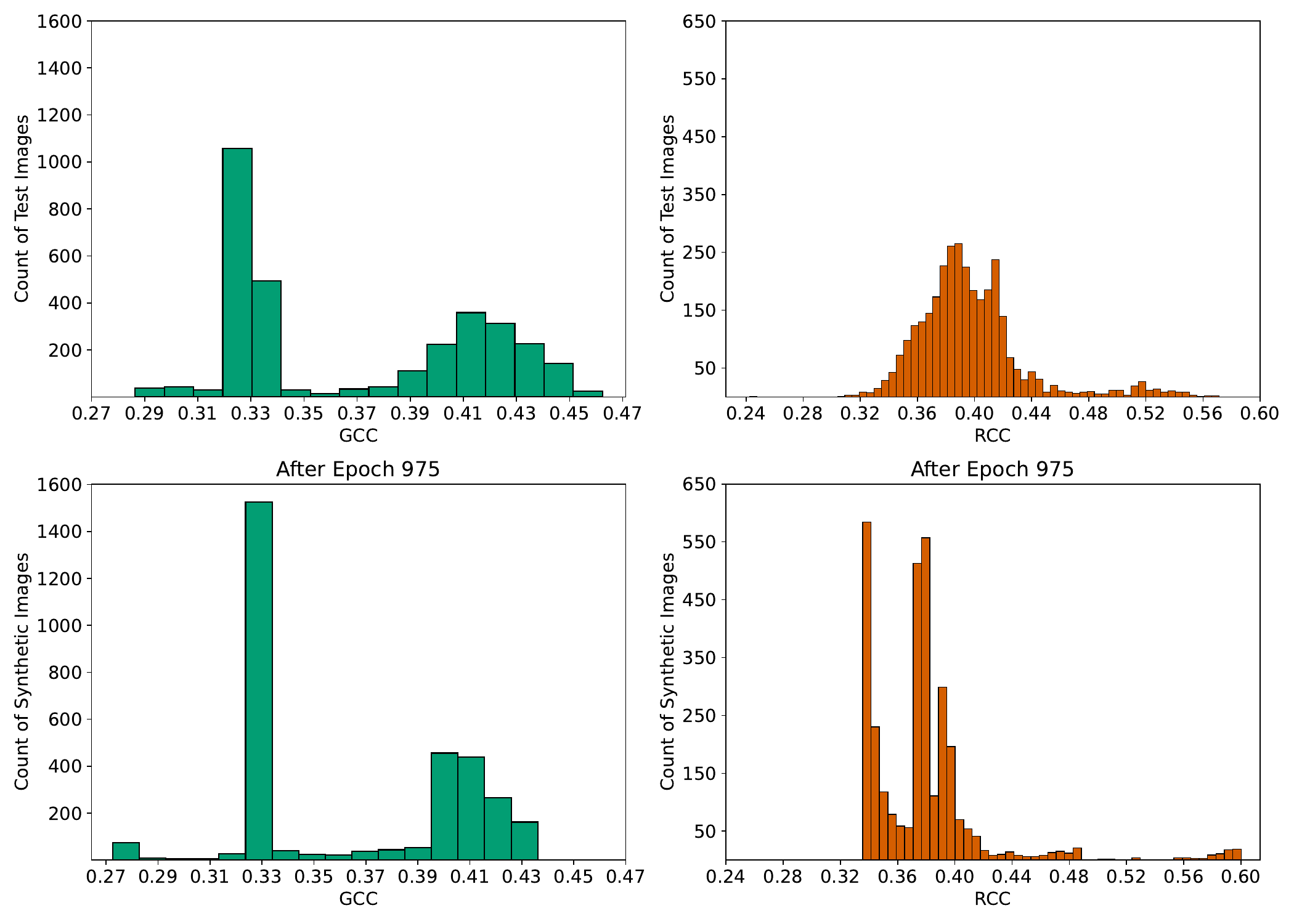}
					\end{center}
			}}
			\subcaption{Harvard Forest}
		\end{subfigure}
		\par\medskip
		\begin{subfigure}[c]{0.85\linewidth}
			\fbox{
				\parbox[][0.40\textheight][c]{0.95\linewidth}{
					\begin{center}
						\includegraphics[width=\linewidth, height=0.40\textheight]{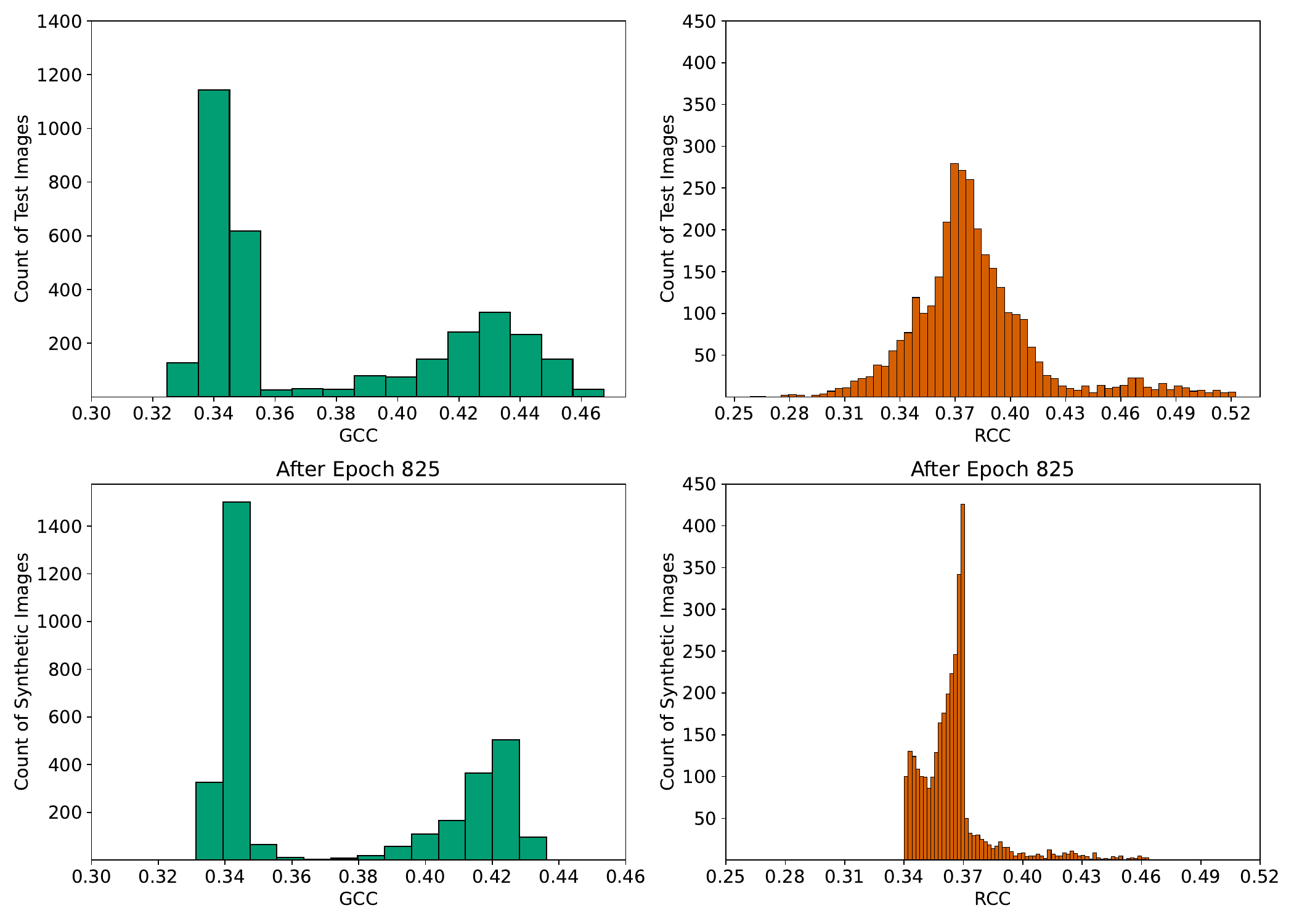}
					\end{center}
			}}
			\subcaption{Bartlett Experimental Forest}
		\end{subfigure}
		\caption{GCC and RCC distribution across test images and corresponding synthetic images over the ``DB\_1000'' ROI.}
		\label{fig:GCC_RCC_Comparison}
	\end{figure}

	A comparative study of GCC and RCC values is conducted between the test images and the synthetic images (Figure \ref{fig:GCC_RCC_Comparison}). We observe that the degree of similarity between the GCC distributions of the test and synthetic images is higher than that of the corresponding RCC distributions. For instance, in Harvard Forest, the highest and second-highest number of test images correspond to GCC values around 0.33 and 0.41, respectively, and the synthetic images exhibit similar characteristics. In case of RCC distribution, the peak count of test images occurs around an RCC value of 0.38, whereas the RCC values of the synthetic images are mostly around 0.34 and 0.38. This observation highlights the success of conditional aspect of our GAN architecture with respect to GCC. 
	
	However, our objective also includes predicting other phenotypic attributes like RCC from the synthetic images. From this perspective, it can be observed that the range of predicted RCC values of synthetic images effectively covers the most frequent RCC values of the test images. Quantitatively, the RMSPE of GCC and RCC values for the generated images, when compared against the ground-truth GCC and RCC values of the test images, are 2.1\% and 9.5\%, respectively, for Harvard Forest, and 2.1\% and 8.4\%, respectively, for Bartlett Forest.   
	
	\subsubsection{Evaluating Efficacy of our GAN model}
	
\begin{figure}[!h]
		\centering
		\begin{subfigure}[c]{0.48\textwidth}
			\centering
			\fbox{
				\parbox[][0.18\textheight][t]{0.95\linewidth}{
					\includegraphics[width=\linewidth, height=0.15\textheight]{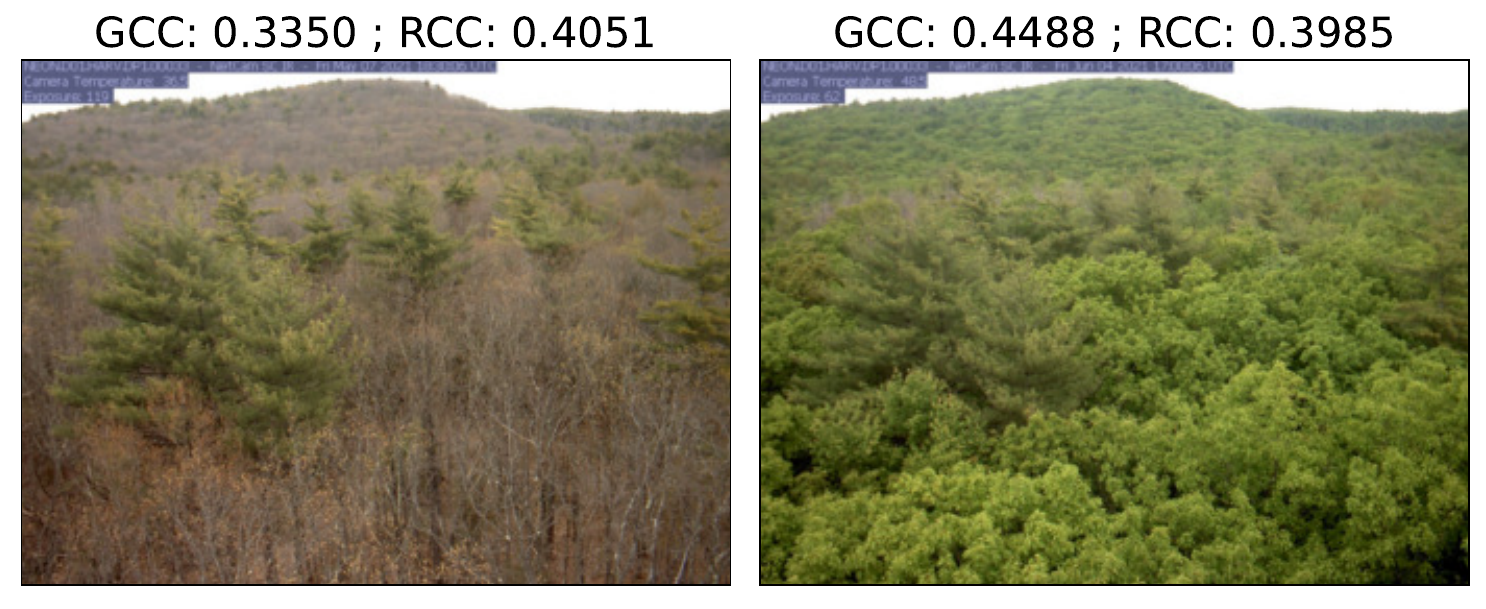}
			}}
			\subcaption{Sample images from test dataset}
		\end{subfigure}
		\medskip
		\begin{subfigure}[c]{0.48\textwidth}
			\centering
			\fbox{
				\parbox[][0.18\textheight][c]{0.95\linewidth}{
					\includegraphics[width=\linewidth, height=0.18\textheight]{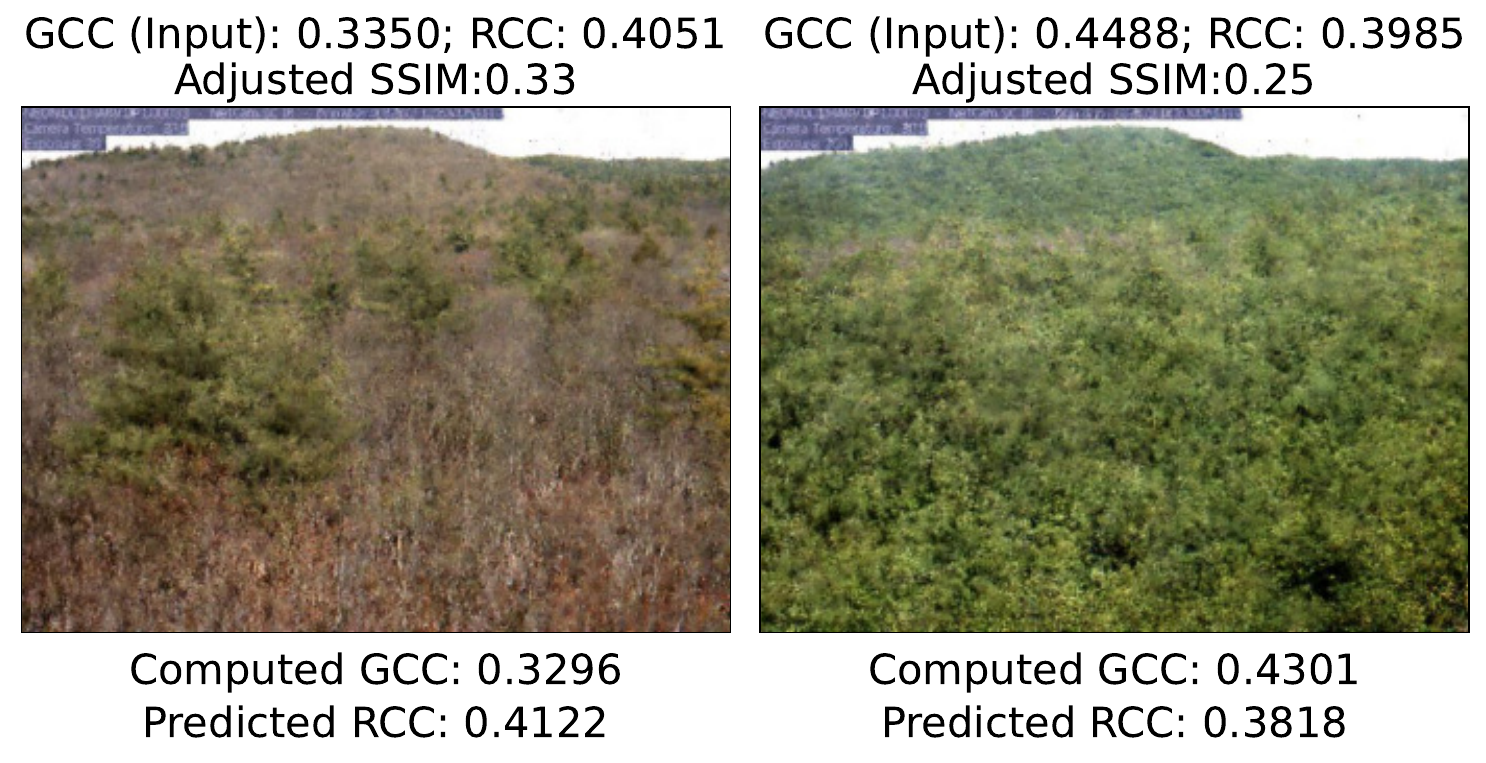}
			}}
			\subcaption{Synthetic images after epoch $975$}
		\end{subfigure}
		\par
		\begin{center}
			\parbox[t][0.5cm][c]{0.78\textwidth}{\caption{Sample test images with GCC values not being used in training and corresponding synthetic images for Harvard Forest depicting the ability of our GAN model to generate synthetic images given the GCC value within the range used in training. GCC and RCC correspond to the ``DB\_1000'' ROI indicated on the right.}\label{fig:continuety}	
			}
			\hspace{1pt}
			\parbox[t][0.5cm][c]{0.15\textwidth}{\includegraphics[width=0.15\textwidth]{HARV_ROI_DB_1000-eps-converted-to.pdf}}
		\end{center}
	\end{figure}

	As previously mentioned, our goal is to develop a GAN model conditioned on a continuous attribute over a specific portion of the image. To evaluate this, we select only those test images corresponding to GCC values that were not used during model (discriminator) training (i.e., GCC values not present in the train dataset). For Harvard Forest, we identify 97 such test images involving 79 adjusted-GCC values. Examples of the original test images with the corresponding generated images are presented in Figure \ref{fig:continuety}. The RMSPE of GCC and RCC for these 97 synthetic images are 5\% and 9.8\%, respectively. In case of Bartlett Forest, we identify $42$ such test images covering $34$ unique adjusted-GCC values, and the RMSPE of GCC and RCC across these images are 4.6\% and 6\%, respectively.        
	
	\subsubsection{Analyzing Significance of Proposed Work}
	
\begin{figure}[!h]
		\centering
		\begin{subfigure}[c]{0.48\textwidth}
			\centering
			\fbox{
				\parbox[][0.18\textheight][t]{0.95\linewidth}{
					
					\includegraphics[width=\linewidth, height=0.15\textheight]{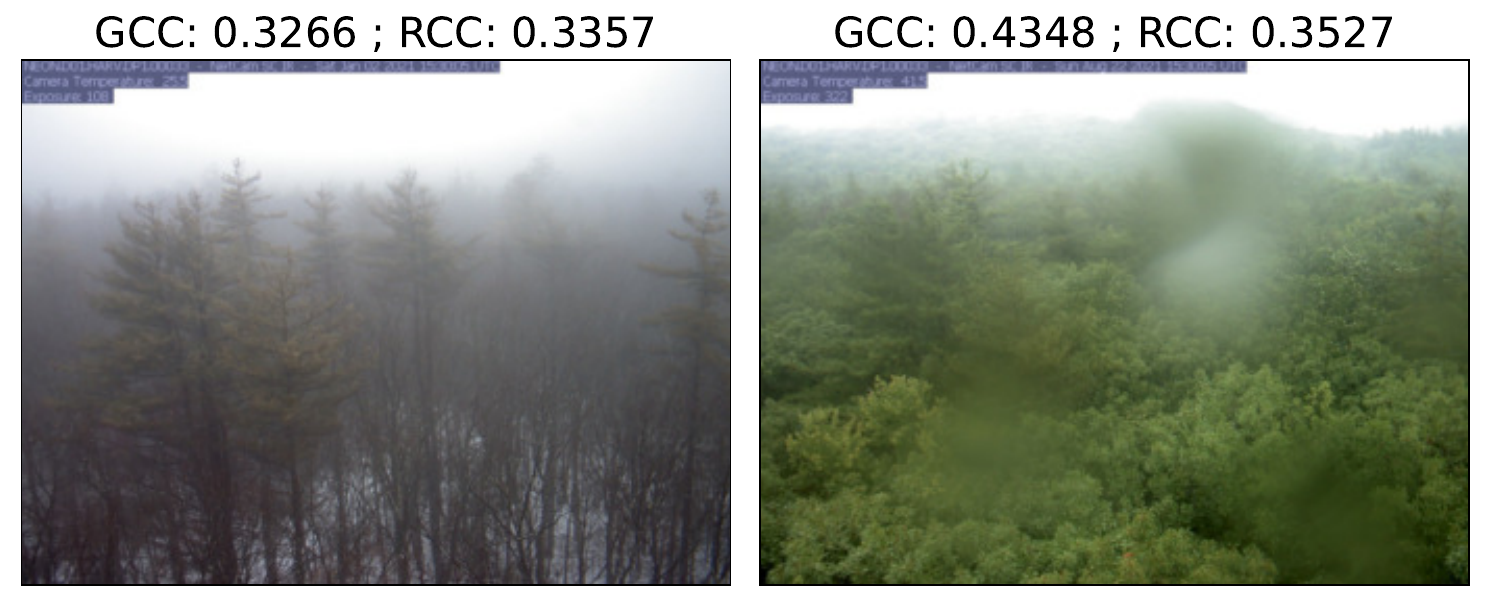}
			}}
			\subcaption{Sample images from test dataset}
		\end{subfigure}
		\bigskip
		\begin{subfigure}[c]{0.48\textwidth}
			\centering
			\fbox{
				\parbox[][0.18\textheight][c]{0.95\linewidth}{
					\begin{center}
						\includegraphics[width=\linewidth, height=0.18\textheight]{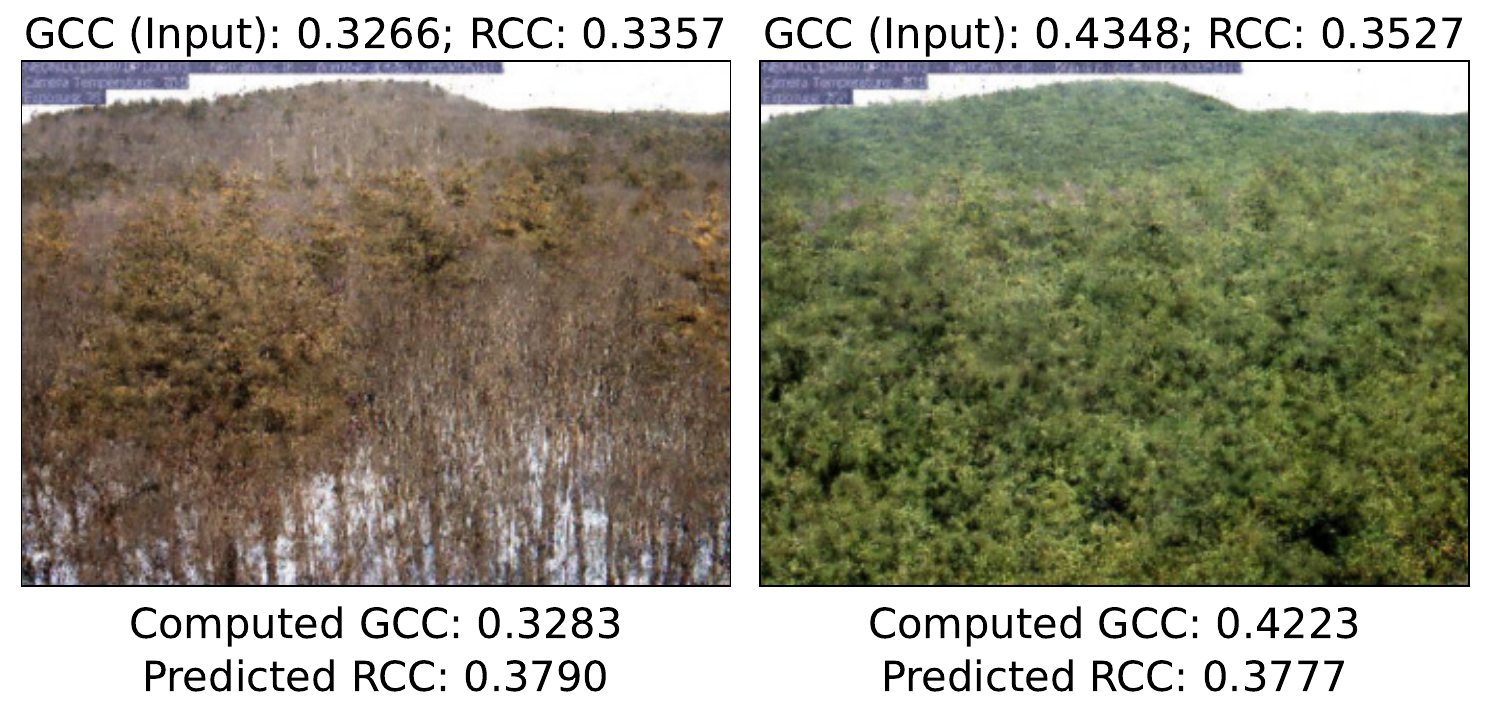}
					\end{center}
			}}
			\subcaption{Synthetic images after epoch $975$}
		\end{subfigure}
		\par
		\begin{center}
			\parbox[c][0.4cm][c]{0.78\textwidth}{\caption{Sample blurred test images and corresponding synthetic images for Harvard Forest illustrating the potential of our GAN model to better portray the appearance of the forest based on a greenness value (GCC and RCC correspond to the ROI ``DB\_1000'' indicated on the right).}\label{fig:significance}	
			}
			\hspace{1pt}
			\parbox[c][0.4cm][c]{0.15\textwidth}{\includegraphics[width=0.15\textwidth]{HARV_ROI_DB_1000-eps-converted-to.pdf}}
		\end{center}
		
	\end{figure}

	A kind of blurriness is detected across PhenoCam images for certain GCC values. This work also aims to improve the visual quality of forest site images based on a greenness value. Therefore, we consider blurred sample images from the test dataset of Harvard Forest and generate synthetic images using our GAN model, as shown in Figure \ref{fig:significance}. The results indicate that the generated images can improve the visualization of forest site appearances in such cases. Moreover, plant biologists could leverage these synthetic images to gain a better understanding of other phenotypic attributes.
	
	\subsubsection{Ablation Study on Self-Attention Modules}
	\label{subsubsec:ablation}
\begin{figure}[!h]
		\centering
		\fbox{
			\parbox[][0.3\textheight][c]{0.98\linewidth}{
				\includegraphics[width=0.98\textwidth, height=0.3\textheight]{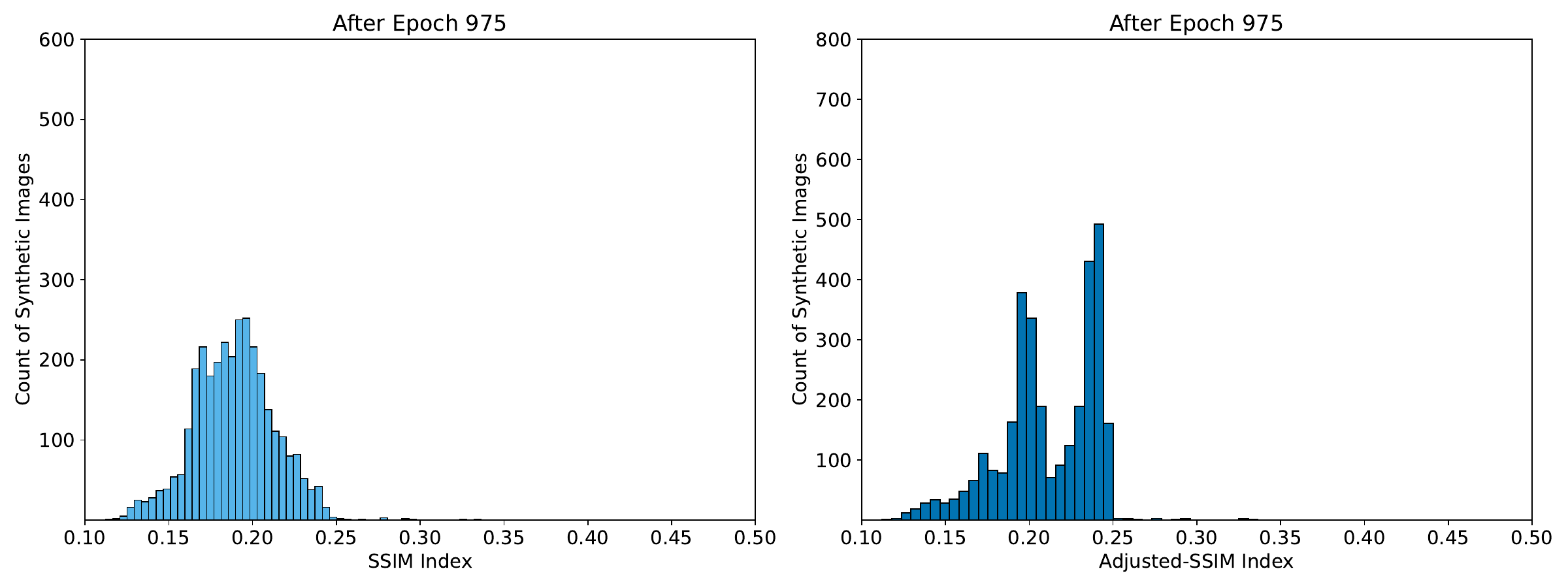}
		}}
		\caption{SSIM index of synthetic images against test images for Harvard Forest after training without self-attention modules. Comparison with Figure \ref{fig:SSIM} shows significant improvement of SSIM scores after incorporating self-attention modules into our GAN model.}
		\label{fig:SSIM_HARV_Without_SA}
	\end{figure}

	To evaluate the impact of self-attention modules on the performance of our GAN architecture, we train the GAN model for the Harvard Forest without incorporating self-attention modules. The similarity scores between the test images and their corresponding generated images for the model trained without self-attention are shown in Figure \ref{fig:SSIM_HARV_Without_SA}. A comparison with the similarity scores obtained with self-attention modules for Harvard Forest (Figure \ref{fig:SSIM}) reveals that the inclusion of self-attention modules improves the quality of the generated images in terms of both SSIM and adjusted-SSIM indices. Specifically, only 0.06\% and 0.09\% of the synthetic images for Harvard Forest, respectively, achieve SSIM and adjusted-SSIM indices greater than 0.3 (the lower bound of the benchmark described in Section \ref{subsubsec:quality}) without self-attention modules. In contrast, with self-attention modules, 21\% and 43.5\% of the synthetic images from Harvard Forest surpass this threshold, respectively. In addition, the FID score for synthetic images generated without self-attention modules from Harvard Forest is 115.1, whereas the inclusion of self-attention modules improves the FID score to 82.1. This demonstrates the effectiveness of self-attention modules in enhancing the performance of our GAN model.       
	
	\subsubsection{Significance of ROI in Proposed Method}
	
	To analyze the significance of providing ROI image as input to the GAN model for generating synthetic images that satisfy the given GCC value over the ROI, we conduct an alternative experiment. In this setup, only the GCC values of the ROI are provided as auxiliary information to the GAN model (both generator and discriminator), omitting the ROI image as input. Under this consideration, the method does not incorporate the masking mechanism to condition GCC over the ROI; instead, the input GCC values are conditioned on the whole image with an assumption that the generator can inherently produce synthetic images satisfying the given GCC value over the ROI. Using the same GAN architecture (Figure \ref{fig:GAN_Architecture}), we train the model on the Harvard Forest dataset (GCC values correspond to the ``DB\_1000'' ROI), by excluding the ROI image as input to the model but using the same training process described in Section \ref{sec:Methodologies}. 
	
\begin{figure}[!h]
	\begin{subfigure}[b]{\textwidth}
		\fbox{
			\parbox[][0.16\textheight][c]{0.95\linewidth}{
				
				\begin{center}
					\includegraphics[width=\linewidth, height=0.16\textheight]{HARV_synthetic_image-eps-converted-to.pdf}
				\end{center}
		}}
		\subcaption{With ROI image}
	\end{subfigure}
	\par\bigskip
	
	\begin{subfigure}[b]{\textwidth}
		\fbox{
			\parbox[][0.16\textheight][c]{0.95\linewidth}{
				
				\begin{center}
					
					\includegraphics[width=\linewidth, height=0.16\textheight]{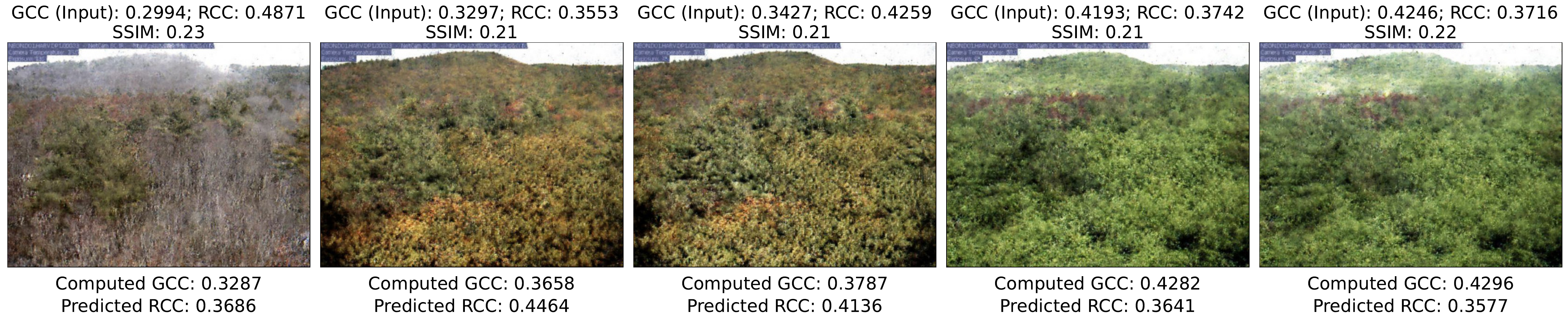}
				\end{center}
		}}
		\subcaption{Without ROI image}
	\end{subfigure}
	\par
	\begin{center}
		\parbox[c][0.5cm][c]{0.8\textwidth}{\caption{Synthetic images generated after training the GAN model with and without ROI information. This is for the Harvard Forest site. Notice the reduction of SSIM scores as well as GCC and RCC values of synthetic images, which underscores the need for the ROI information.}\label{fig:HARV_image_wo_roi}	
		}
		\hspace{1pt}
		\parbox[c][0.5cm][c]{0.15\textwidth}{\includegraphics[width=0.15\textwidth]{HARV_ROI_DB_1000-eps-converted-to.pdf}}
	\end{center}
	
\end{figure}

	We observe significant performance degradation in the absence of ROI information. Only 0.03\% of the synthetic images achieve SSIM and adjusted-SSIM scores greater than 0.3 (the lower bound of the benchmark described in Section \ref{subsubsec:quality}), compared to 21\% and 43.5\% with ROI information (Section \ref{subsubsec:quality}), respectively. Further, the FID score of synthetic images increases from 82.1 (with ROI information) to 118.2 (without ROI information), indicating a decline in the quality of generated images. Additionally, the RMSPE of synthetic images increases from 2.1\% (GCC) and 9.5\% (RCC) with ROI information (Section \ref{subsubsec:gcc_rcc}) to 8.2\% (GCC) and 14.3\% (RCC) without ROI information. This analysis underscores the importance of incorporating ROI as part of the input to our proposed GAN model. Figure \ref{fig:HARV_image_wo_roi} presents samples of synthetic images generated for the same GCC values with and without ROI information.
	
%
%

	\subsubsection{Investigating Limitations of Proposed Method}
	
\begin{figure}[!h]		
		\centering
		\begin{subfigure}[t]{0.99\linewidth}
			\centering
			\fbox{
				\parbox[][0.3\textheight][c]{0.98\linewidth}{
					\begin{center}
						\includegraphics[width=\linewidth, height=0.28\textheight]{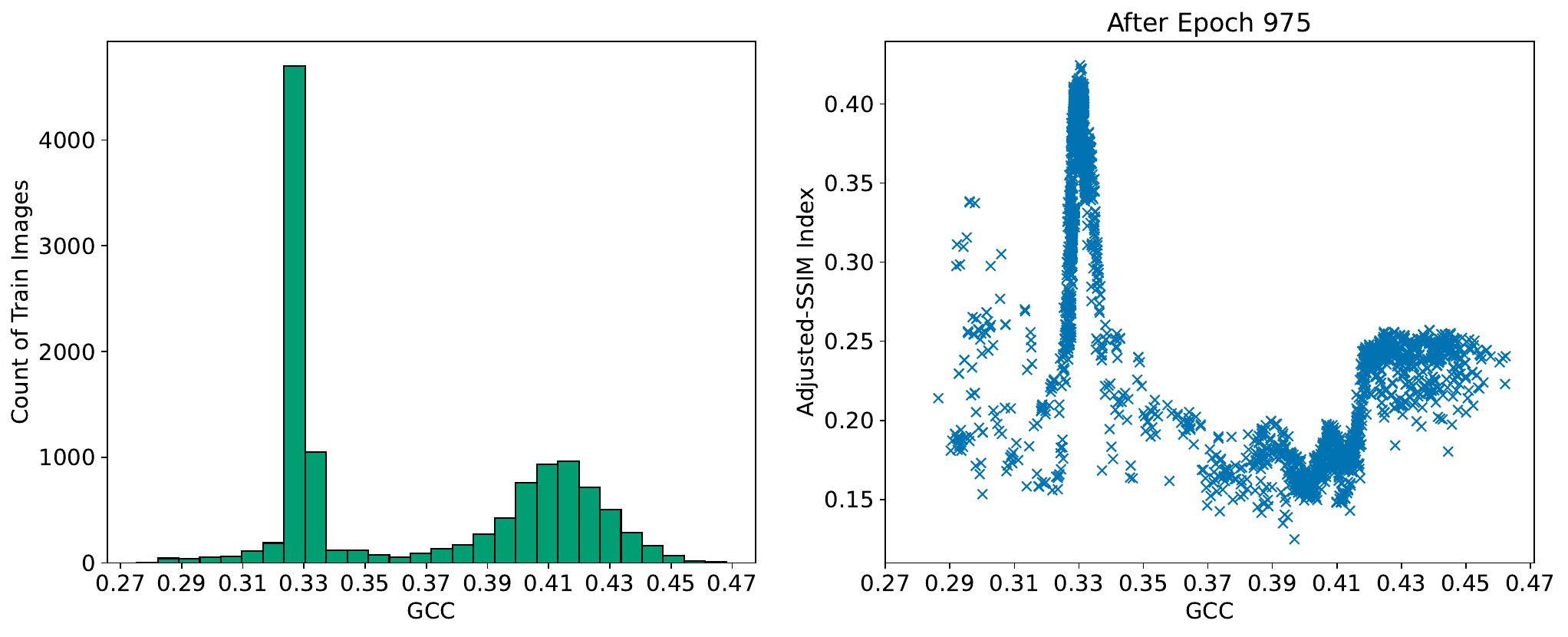}
					\end{center}
			}}
			\subcaption{Harvard Forest}
		\end{subfigure}
		\par\medskip
		\begin{subfigure}[t]{0.99\linewidth}
			\centering
			\fbox{
				\parbox[][0.3\textheight][c]{0.98\linewidth}{
					\begin{center}
						\includegraphics[width=\linewidth, height=0.3\textheight]{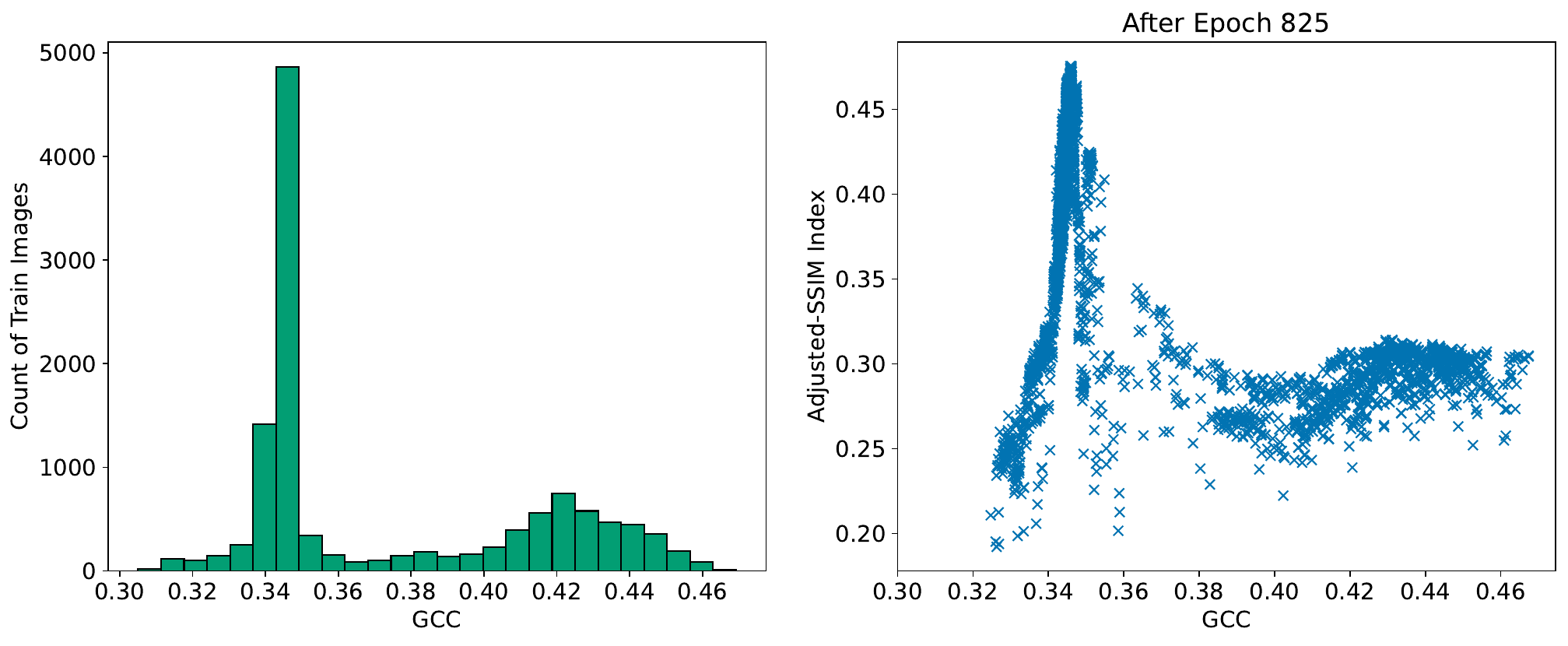}
					\end{center}
			}}
			\subcaption{Bartlett Experimental Forest}
		\end{subfigure}
		\caption{GCC distribution across training images (left) and adjusted-SSIM score distribution across synthetic images (right) depicting the impact on model performance due to limited training data. The adjusted-SSIM index is the maximum score obtained for the synthetic image after comparing it with all the test images corresponding to the given GCC value.}
		\label{fig:SSIM_GCC_distribution}		
	\end{figure}

	We also examine the limitations of our proposed GAN model. The adjusted-SSIM scores of synthetic images generated by our method are comparatively higher over certain GCC values than others within the specified range for both forest sites (right side of Figure \ref{fig:SSIM_GCC_distribution}). These discrepancies correlate with the number of training images available for different GCC values (left side of Figure \ref{fig:SSIM_GCC_distribution}). This observation indicates that the performance of our model is influenced by the limited availability of the training images for certain GCC values. Consequently, our proposed method struggles to consistently generate synthetic images with conformable quality across the entire specified range of greenness, particularly in regions where the training data is sparse.
	
\begin{figure}[!h]
		\centering
		\fbox{
			\parbox[][0.16\textheight][c]{0.98\linewidth}{
				
				\begin{center}
					\includegraphics[width=\linewidth, height=0.16\textheight]{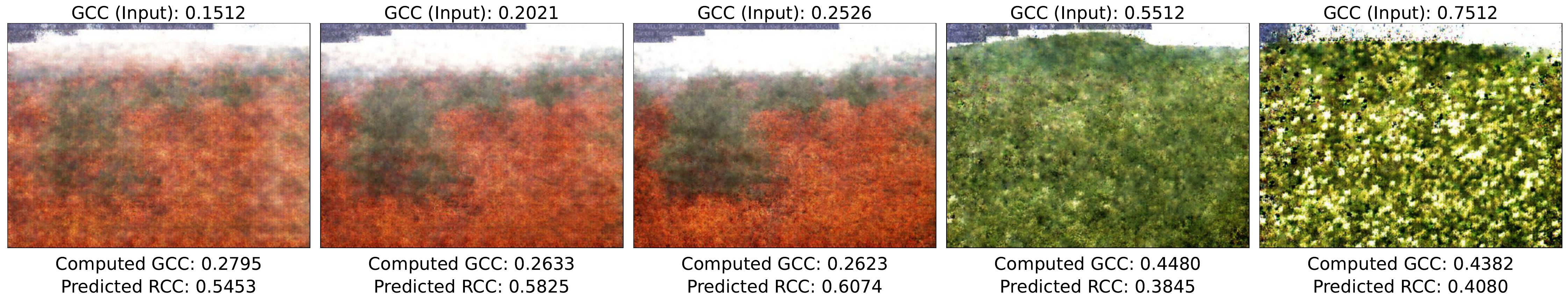}
				\end{center}
		}}
		\par
		\bigskip
		\begin{center}			
			\parbox[t][1.3cm][c]{0.8\textwidth}{\caption{Impact of providing GCC values outside the range specified during training. These examples show synthetic images of Harvard Forest corresponding to GCC values outside the specified range [0.27,0.47] for the ``DB\_1000'' ROI. When GCC values lower than the range are provided, the generated images tend to be red, whereas input GCC values greater than the range result in synthetic images that tend to be green, indicating their proximity to the lower and upper bounds of the range, respectively.}\label{fig:HARV_image_outside_range}	
			}
			\hspace{1pt}
			\parbox[c][0.5cm][c]{0.15\textwidth}{\includegraphics[width=0.15\textwidth]{HARV_ROI_DB_1000-eps-converted-to.pdf}}
		\end{center}
		
	\end{figure}

	Next, we examine the model output corresponding to GCC values outside the specified range used during training. Figure \ref{fig:HARV_image_outside_range} displays examples of synthetic images from Harvard Forest corresponding to GCC values that fall outside the range used during training for the ``DB\_1000'' ROI. Since no real images exist for these GCC values, we cannot assess the quality of these synthetic images using SSIM scores. Visually, these generated images exhibit blurriness, and the quality of the images deteriorates as the GCC value moves farther away from the specified range. However, when provided with GCC values lower than the specified range as input to the model, the generated images tend to be red in color, and the computed GCC values are closer to the lower bound of the range. Conversely, when input GCC values are greater than the specified range, the model generates images that are likely to be green in color, with computed GCC values approaching the upper bound of the range. 
	
	\subsection{Cross-site Training: Harvard to Bartlett Forest and Vice-Versa}
	
	We conduct a study to evaluate the generalizability of our model, specifically to determine whether a model trained on one forest site can be extended to other sites using a small amount of data and fewer training iterations. Importantly, a model trained on one forest site cannot be directly generalized to other sites without fine-tuning due to distinct morphological structures and unique ROIs describing the vegetation type of each site (Figure \ref{fig:real_images}). To address this, the model trained on the Harvard Forest dataset for 975 epochs (as described in Section \ref{subsec:individual_training}), referred to as the base model, is fine-tuned with 50\% of the training data from Bartlett Forest for 100 epochs using training parameters derived from Bartlett data. 
	
\begin{figure}[h!]
		\begin{subfigure}[b]{\textwidth}
			\centering
			\fbox{
				\parbox[][0.13\textheight][c]{0.95\linewidth}{
					\centering
					\includegraphics[width=0.95\textwidth, height=0.13\textheight]{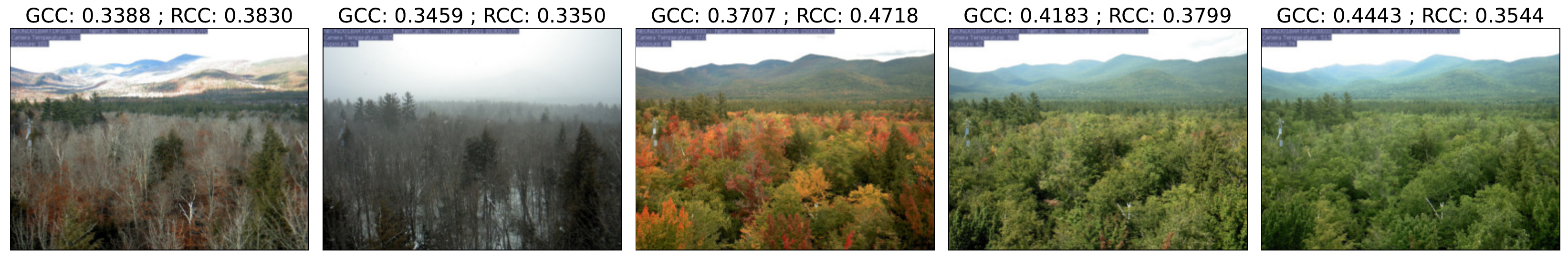}
			}}
			\subcaption{Sample Images from Test Dataset}
			
		\end{subfigure}
		\par\medskip
		\begin{subfigure}[b]{\textwidth}
			\centering
			\fbox{
				\parbox[][0.16\textheight][c]{0.95\linewidth}{
					\centering
					\includegraphics[width=0.95\textwidth, height=0.16\textheight]{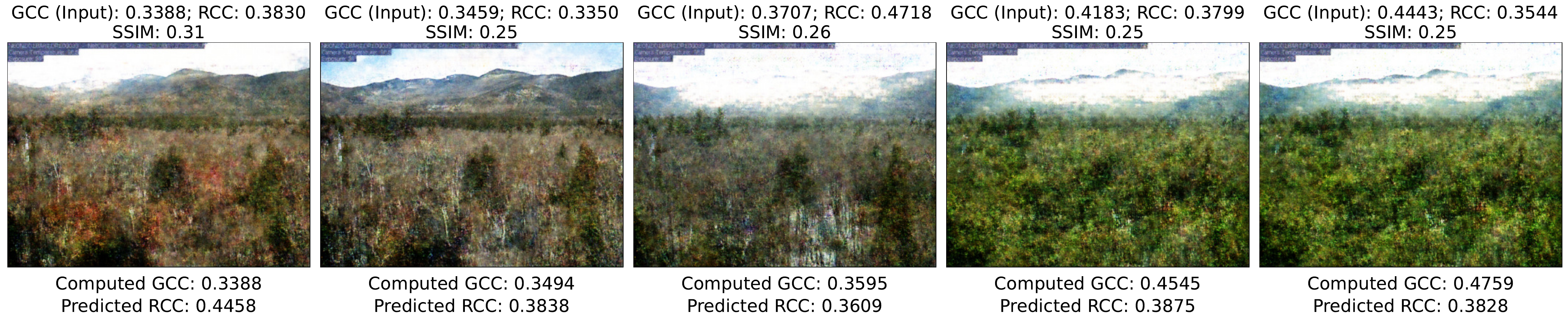}
			}}
			\subcaption{Case 1: Synthetic images after epoch 100 (Trained from scratch with the full Bartlett Forest training dataset)}
			
		\end{subfigure}
		\par\medskip
		\begin{subfigure}[b]{\textwidth}
			\centering
			\fbox{
				\parbox[][0.16\textheight][c]{0.95\linewidth}{
					\centering
					\includegraphics[width=0.95\textwidth, height=0.16\textheight]{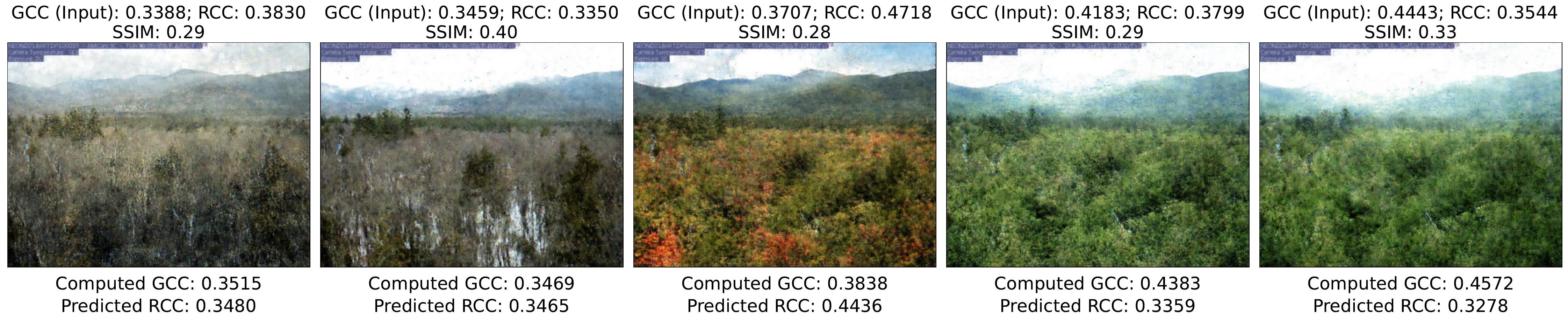}
			}}
			\subcaption{Case 2: Synthetic images after epoch 100 (Updated the Harvard Forest model using 50\% of the Bartlett Forest training data)}
		\end{subfigure}
		\begin{center}
			\par
			\parbox[c][0.5cm][c]{0.8\textwidth}{\caption{Cross-site experiment: Sample test images and synthetic images for Bartlett Experimental Forest (SSIM indicates the similarity score of synthetic image with the corresponding test image. GCC and RCC correspond to the ``DB\_1000'' ROI indicated on the right).}\label{fig:BART_image_cross_site}	
			}
			\hspace{1pt}
			\parbox[c][0.5cm][c]{0.15\textwidth}{\includegraphics[width=0.15\textwidth]{BART_ROI_DB_1000-eps-converted-to.pdf}}
		\end{center}
	\end{figure}

\begin{figure}[h!]			
	\centering
	\begin{subfigure}[t]{0.98\linewidth}
		\centering
		\fbox{
			\parbox[][0.28\textheight][c]{0.95\linewidth}{
				\begin{center}
					\includegraphics[width=\linewidth, height=0.28\textheight]{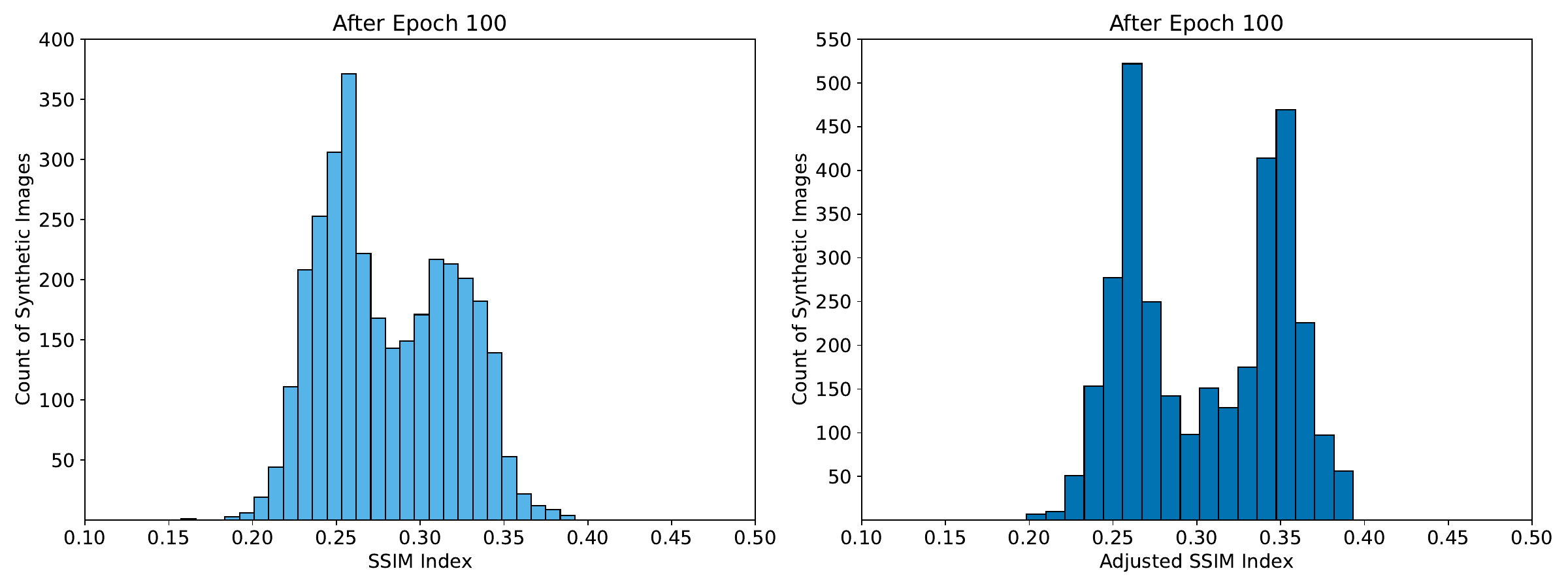}
				\end{center}
		}}
		\subcaption{Case 1: Trained from scratch with the full Bartlett Forest training dataset}
	\end{subfigure}
	\par
	\medskip
	\begin{subfigure}[t]{0.98\linewidth}
		\centering
		\fbox{
			\parbox[][0.28\textheight][c]{0.95\linewidth}{
				\begin{center}
					\includegraphics[width=\linewidth, height=0.28\textheight]{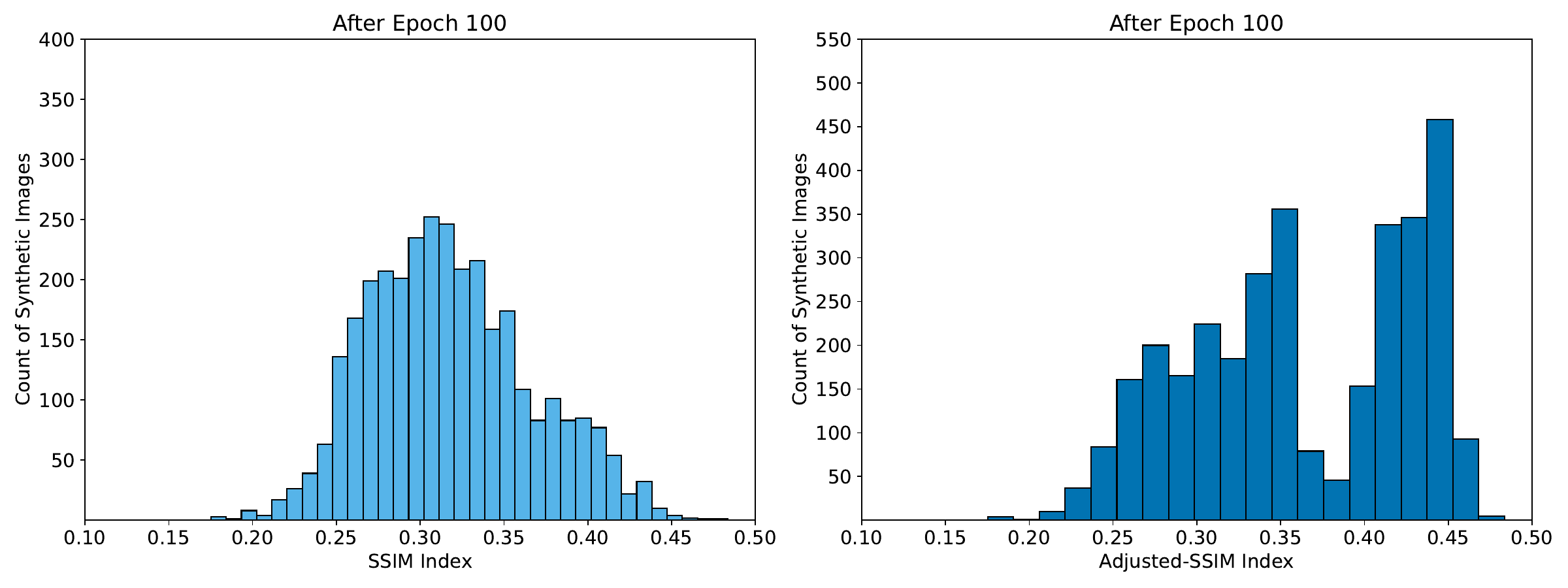}
				\end{center}
		}}
		\subcaption{Case 2: Updated the Harvard Forest model with 50\% of the Bartlett Forest training data}
	\end{subfigure}
	\caption{Cross-site experiment: Comparison of SSIM index for synthetic images against the test dataset for Bartlett Experimental Forest.}
	\label{fig:BART_cross_site_SSIM}
\end{figure}

	Our findings indicate that the fine-tuned model effectively adapts to generate synthetic images for the target site. We perform a comparative analysis between (1) a model trained from scratch on the full Bartlett Forest dataset for 100 epochs, and (2) the cross-site fine-tuned model. Figure \ref{fig:BART_image_cross_site} shows sample test images from Bartlett Forest alongside their corresponding synthetic images for both cases. In Case (1), the RMSPEs of synthetic images are 4.8\% (GCC) and 12.6\% (RCC), whereas in Case (2), RMSPEs improve to 3\% (GCC) and 12.2\% (RCC). The SSIM and adjusted-SSIM scores for both cases are presented in Figure \ref{fig:BART_cross_site_SSIM}. Additionally, the FID score of synthetic images in Case (1) is 193.7, while in Case (2), it significantly improves to 127.9. Remarkably, this FID score is comparable to that achieved by the original Bartlett Forest model (Section \ref{subsec:individual_training}), which was trained on the full Bartlett dataset for 875 epochs, highlighting the effectiveness of the adaptation process. However, the FID score of the images generated by the cross-site fine-tuned model for Bartlett Forest is lower in quality than that of the base model trained on Harvard Forest dataset (82.1).
	
	Next, we perform a similar fine-tuning experiment to adapt the Bartlett Forest model to generate images for Harvard Forest. To achieve this, the Bartlett Forest model, trained for 825 epochs (as described in Section \ref{subsec:individual_training}), is fine-tuned with 50\% of the training data from Harvard Forest for 100 epochs using training parameters derived from Harvard data. A comparison between (1) the model trained from scratch on the full Harvard Forest dataset for 100 epochs and (2) the cross-site fine-tuned model reveals that the RMSPEs of synthetic images are 4.2\% (GCC) and 8.5\% (RCC) in Case (1), compared to 2.9\% (GCC) and 10.1\% (RCC) in Case (2). The FID score of synthetic images in Case (1) is 111.3, whereas in Case (2), it increases to 159.5. Here, the FID score of the cross-site fine-tuned model is higher, indicating lower quality images compared to the original Harvard Forest model (FID score: 82.1). This degradation may potentially be attributed to the relatively higher FID score of the base model trained on Bartlett Forest dataset (127.1). Nevertheless, this increase in the FID score of the cross-site fine-tuned model from the base model aligns with the trend observed in the previous case, reflecting a consistent trade-off in cross-site adaptation.
	
	\subsection{Scalability to Other Vegetation Type}
	
\begin{figure}[!h]
		\begin{subfigure}[b]{\textwidth}
			\centering
			\fbox{
				\parbox[][0.13\textheight][c]{0.95\linewidth}{
					\centering
					\includegraphics[width=0.95\textwidth, height=0.13\textheight]{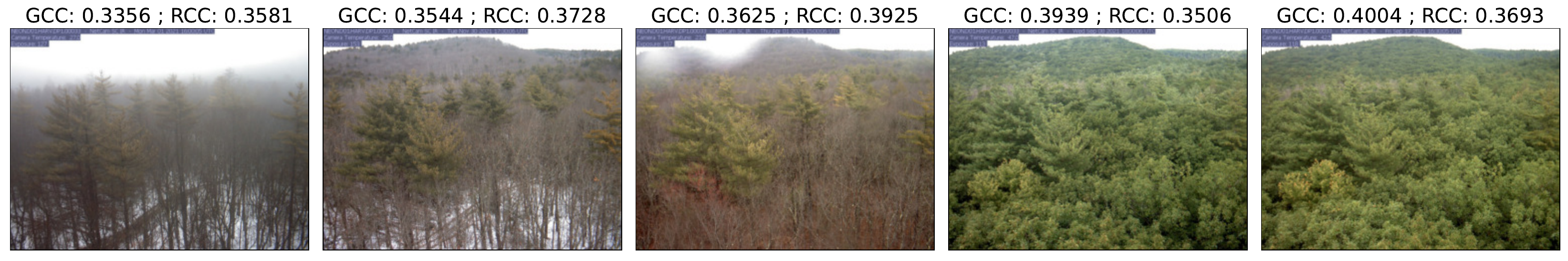}
			}}
			\subcaption{Sample images from test dataset}
			
		\end{subfigure}
		\par\medskip
		\begin{subfigure}[b]{\textwidth}
			\centering
			\fbox{
				\parbox[][0.16\textheight][c]{0.95\linewidth}{
					\centering
					\includegraphics[width=0.95\textwidth, height=0.16\textheight]{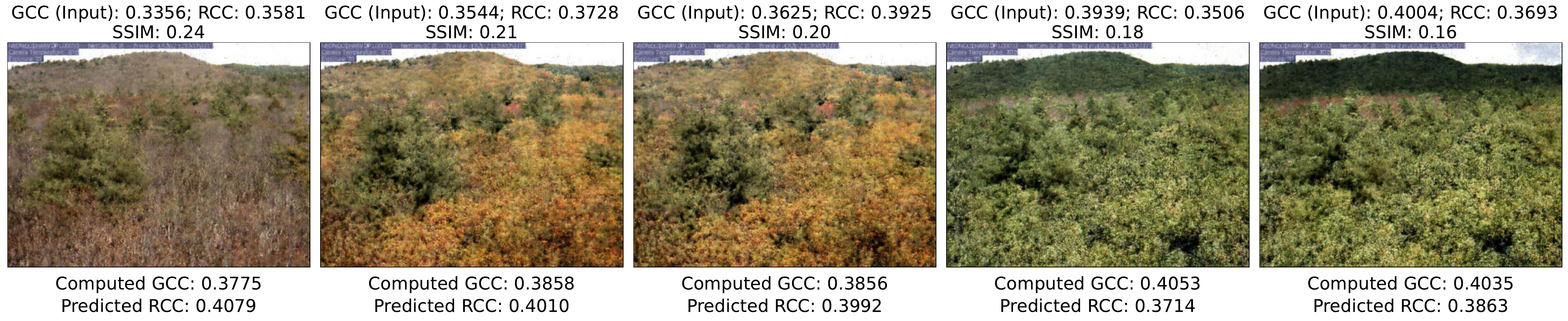}
			}}
			\subcaption{Synthetic images after epoch 975 (Using the Harvard Forest model trained with the ``DB\_1000'' ROI)}
			
		\end{subfigure}
		\par\medskip
		\begin{subfigure}[b]{\textwidth}
			\centering
			\fbox{
				\parbox[][0.16\textheight][c]{0.95\linewidth}{
					\centering
					\includegraphics[width=0.95\textwidth, height=0.16\textheight]{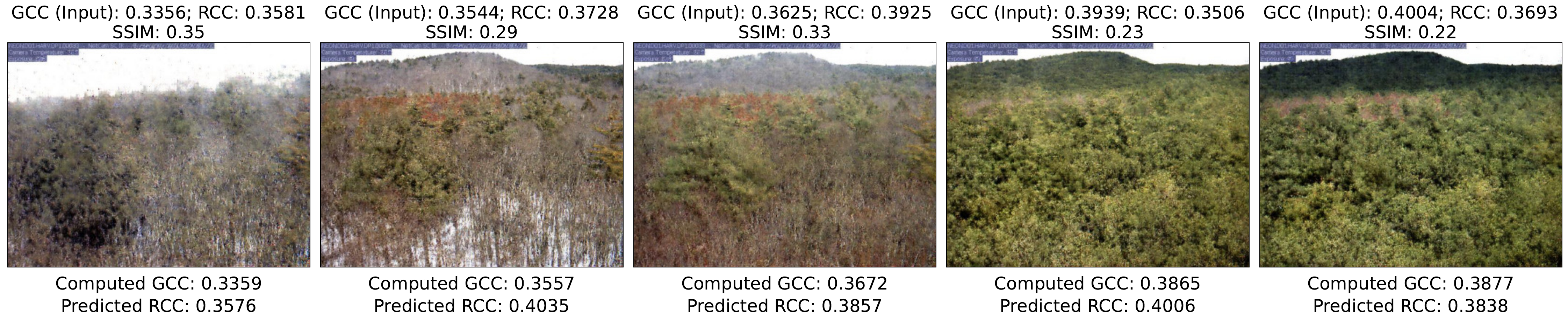}
			}}
			\subcaption{Synthetic images after epoch 25 (After training on top of Harvard Forest model (``DB\_1000'') using 25\% of the ``EN\_1000'' ROI training data)}
			
		\end{subfigure}
		\begin{center}
			\par
			\parbox[c][0.5cm][c]{0.8\textwidth}{\caption{Cross-vegetation experiment: Sample test images and synthetic images for Harvard Forest (SSIM indicates the similarity score of synthetic image with the corresponding test image. GCC and RCC correspond to the ``EN\_1000'' ROI indicated on the right).}\label{fig:HARV_image_cross_vegetation}	
			}
			\hspace{1pt}
			\parbox[c][0.5cm][c]{0.15\textwidth}{\includegraphics[width=0.15\textwidth]{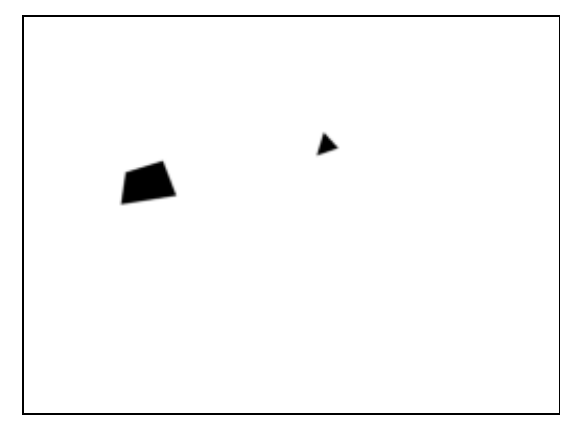}}
		\end{center}
	\end{figure}

	Moreover, we assess the scalability of our GAN model trained on a particular vegetation type to another vegetation type within the same forest site. For this, the model trained on the Harvard Forest dataset with the ``DB\_1000'' ROI, as described in Section \ref{subsec:individual_training}, is first evaluated with another ROI, ``EN\_1000'', within the same forest site. The RMSPE of 6.4\% (GCC) and 7.11\% (RCC) are initially obtained. The model is then further trained for 25 epochs using 25\% of the training data from Harvard Forest for the ``EN\_1000'' ROI, which results in improved RMSPEs of 2.4\% (GCC) and 7.06\% (RCC). Figure \ref{fig:HARV_image_cross_vegetation} shows sample images from the test dataset and the corresponding synthetic images for each of these cases. We observe that both the SSIM index and GCC improve after extending the ``DB\_1000'' model using just 25\% of the training data from the new ROI, ``EN\_1000''. Additionally, the FID score of synthetic images from Harvard Forest improves to 75.2 from 82.1. These results further demonstrate that our proposed model is fairly robust to the size and location of the ROI. 
	
	\section{Conclusion}
	\label{sec:conclusion}
	
	In this work, we present a novel GAN architecture for synthesizing forestry images that satisfy a specific phenotypic attribute, viz., greenness index over the ROI of an image. Experiments conducted on the PhenoCam dataset indicates that the synthetic images generated by our GAN model can be utilized to (a) visualize the appearance of a forest site based on the greenness value, and (b) predict other phenotypic attributes (e.g., redness index) that were not involved during image synthesis. The SSIM scores between the generated and real images were found to be analogous to the SSIM scores between real images, thereby substantiating the quality of the generated images. Further, the proposed model is capable of producing synthetic images with favorable FID scores, as well as a diverse set of images corresponding to a specific GCC value. It can also generate forestry images corresponding to GCC values not used during training but within the defined range for the forest site. Moreover, we demonstrated that our GAN model trained on one forest site can be fine-tuned to generate images for other forest sites, establishing the generalization capability of the model. In addition, the model is scalable to other vegetation types within the same forest site in an efficient manner.
	
	From a broader perspective, this work aims to advance the study on image generation by identifying patterns in images that do not have distinct morphological structure. Instead, our model automatically learned the phenomenon of green-up and green-down based on the colors and textures of images. Additionally, we applied conditioning of a continuous attribute on a certain portion of the image (ROI), which provided control over the image generation process. Moreover, this technique could be leveraged to visualize forestry based on different plant phenotypes (e.g., canopy cover) in the context of various environmental parameters (e.g., temperature, precipitation). 
	
	However, due to the limited size of the training dataset and asymmetric distribution of GCC values across the training images, the proposed model was unable to generate high-quality images for certain GCC values. It must also be noted that due to computational and time constraints, the size of the generated images was set to be smaller than that of the original PhenoCam images. Additionally, our proposed model only captures the phenotypic relationship patterns in a forest site without considering changes in environmental parameters over the years. 
	
	Currently, we are working to further improve the quality of the generated forestry images by using stable diffusion models \cite{DN2021}. This approach could also be extended to other forest sites within different NEON domains. Additionally, other phenological and phenotypical information (e.g., LAI, canopy cover) could be extracted from the synthetic images. In the future, our methodology could potentially be enhanced by accounting for various weather parameters and ecological factors. We believe that the work reported in this paper provides a first step in leveraging generative AI principles from pattern recognition and computer vision for plant phenological research. 
	
	\section*{Acknowledgments}
	This work is supported by National Science Foundation (NSF Award 1940059). We thank all other project members, especially Dr. Bryan Heidorn, Dr. David LeBauer, Dr. Jessica Guo, Dr. Anne Thessen, Dr. Laurel Cooper and Dr. Pankaj Jaiswal. We also thank the reviewers for their valuable suggestions.

	
	
	\bibliographystyle{elsarticle-num} 
	\bibliography{References}
	
	
	%
	%
	%
\end{document}